\documentclass{article}

\usepackage{enumerate}
\usepackage{microtype}
\usepackage{graphicx}
\usepackage{booktabs} % for professional tables
\usepackage{makecell}
\usepackage{graphicx}
\usepackage{pifont}

\usepackage{courier}
\usepackage{multirow}
\usepackage[usenames]{color}
\usepackage{fullpage}
\usepackage{todonotes}
\usepackage{amssymb}
\usepackage{gensymb}
\usepackage{amsfonts}       % blackboard math symbols
\usepackage{enumitem}
\usepackage{wrapfig}
\usepackage{multirow}
\usepackage{algorithmic}
\usepackage[linesnumbered, ruled]{algorithm2e}
\usepackage{amsmath,nccmath}
\usepackage[colorlinks,linkcolor=red,anchorcolor=blue,citecolor=blue]{hyperref}

\usepackage[utf8]{inputenc} % allow utf-8 input
\usepackage[T1]{fontenc}    % use 8-bit T1 fonts
\usepackage{hyperref}       % hyperlinks
\usepackage{url}            % simple URL typesetting
\usepackage{booktabs}       % professional-quality tables
\usepackage{amsfonts}       % blackboard math symbols
\usepackage{nicefrac}       % compact symbols for 1/2, etc.
\usepackage{microtype}      % microtypography
\usepackage{xcolor}         % colors
\usepackage{graphicx}
\usepackage{subfigure}
\usepackage{booktabs}
\usepackage{amsmath}
\usepackage{bbm}
\usepackage{amssymb}
\usepackage{mathrsfs}
\usepackage{multirow}
\usepackage{enumitem}
\usepackage{makecell}
\usepackage{wrapfig}
\usepackage[numbers, compress]{natbib}
\usepackage{listings}

\usepackage{tikz}

\usetikzlibrary{shapes.geometric, calc,arrows.meta,positioning}

\tikzset{
    rect/.style={rectangle, rounded corners, minimum width=1cm, minimum height=1cm,text centered, draw=black, fill=orange!30},
    every label/.style={draw=none},
}
\tikzstyle{arrow} = [thick,<->,>=stealth]

\usepackage{amsthm}
\theoremstyle{definition}
\newtheorem{definition}{Definition}
\newtheorem{property}{Property}

\newcommand{\kai}[1]{{\color{black}#1}}

\makeatletter
\def\blfootnote{\gdef\@thefnmark{}\@footnotetext}
\makeatother

\lstdefinestyle{mystyle}{
  backgroundcolor=\color{white},
  keywordstyle=\color{magenta},
  stringstyle=\color{codepurple},
}

\begin{document}
\title{\LARGE\bf Learning to generate imaginary tasks for improving generalization in meta-learning}
\author{
Yichen Wu\thanks{City University of Hong Kong}
~,~
Long-Kai Huang\thanks{Tencent AI Lab}
~,~
Ying Wei\footnotemark[1] \thanks{Corresponding author}
~,~
% \thanks{Tencent AI Lab; e-mail: {\tt jzhuang@uta.edu}}
}
\date{}
\maketitle

% \author{
%   Yichen Wu\textsuperscript{1,2} , Longkai Huang\textsuperscript{2} , Ying Wei\textsuperscript{1}\thanks{Corresponding Author.} \\
%   City University of Hong Kong\textsuperscript{1} \quad Tencent AI Lab\textsuperscript{2}\\
%   \texttt{yichewu2-c@my.cityu.edu.hk, hlongkai@gmail.com,  yingwei@cityu.edu.hk} \\
% }
% \date{}
% \maketitle

\begin{abstract}
The success of meta-learning on existing benchmarks is predicated on the assumption that the distribution of meta-training tasks covers meta-testing tasks.
Frequent violation of the assumption in applications with either insufficient tasks or a very narrow meta-training task distribution leads to memorization or learner overfitting.
Recent solutions have pursued augmentation of meta-training tasks, while it is still an open question to generate both correct and sufficiently imaginary tasks.
In this paper, we seek an approach that up-samples meta-training tasks from the task representation via a task up-sampling network.
Besides, the resulting approach named Adversarial Task Up-sampling (ATU) suffices to generate tasks that can maximally contribute to the latest meta-learner by maximizing an adversarial loss.
On few-shot sine regression and image classification datasets, we empirically validate the marked improvement of ATU over state-of-the-art task augmentation strategies in the meta-testing performance and also the quality of up-sampled tasks.
\end{abstract}
\label{sec:introduction}
\section{Introduction}
The past few years have seen the burgeoning development of
meta-learning, \emph{a.k.a.} learning to learn, which draws upon the meta-knowledge learned from previous tasks (i.e., \emph{meta-training tasks}) to expedite the learning of novel tasks (i.e., \emph{meta-testing tasks}) with a few examples. 
A sufficient number and diversity of meta-training tasks are pivotal for the generalization capability of the meta-knowledge, so that (1) they cover the true task distribution (i.e., environment~\cite{baxter1997bayesian}) from which~meta-testing~tasks are sampled, discouraging learner overfitting~\cite{rajendran2020meta} and (2) the meta-knowledge empowers fast adaptation via the support set for each task, avoiding memorization overfitting~\cite{yin2020meta}. 
Notwithstanding up to~millions of meta-training tasks in benchmark datasets~\cite{ravi2016optimization,triantafillou2019meta}, real-world applications such as drug discovery~\cite{yao2021improving} and medical image diagnosis~\cite{li2020difficulty} usually have access to only thousands or hundreds of tasks, which puts the meta-knowledge at high risk of learner and memorization~overfitting.

\begin{figure}
    \centering
    \includegraphics[width=\textwidth]{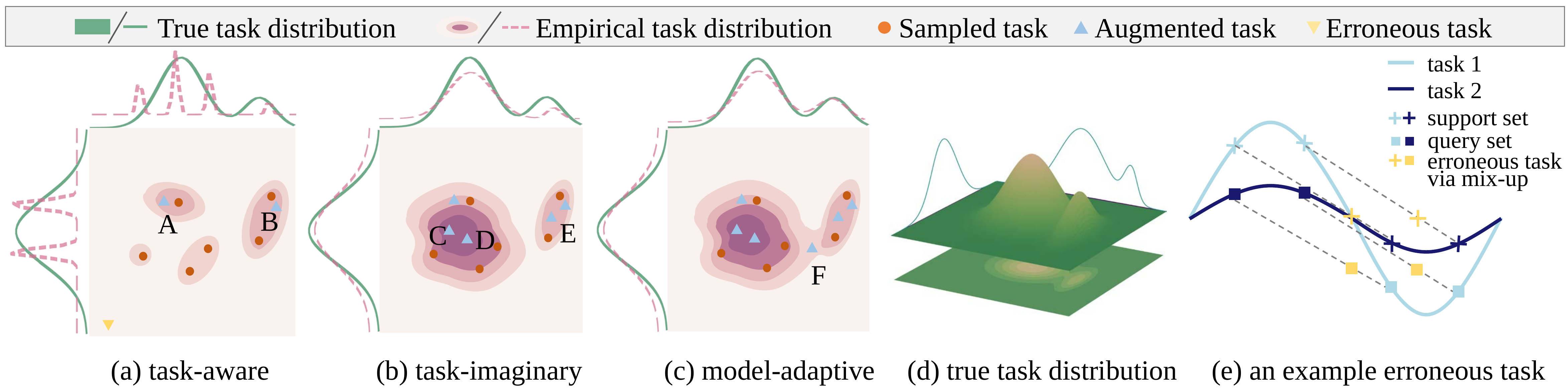}
    \caption{Pictorial illustration of the three characteristics possessed by a qualified task augmentation approach, i.e., being (a) task-aware, (b) task-imaginary, and (c) model-adaptive. (d) shows the true task distribution that task augmentation aims to approximate, and (e) presents an example erroneous task violation of the true task distribution of sinusoidal functions $y=w\sin(x)$ ($w\in[0,2]$).}
    \label{fig:intro}
\end{figure}

While early attempts towards improving the generalization capability of the meta-knowledge revolve around regularization methods that limit the capacity of the meta-knowledge~\cite{jamal2019task,yin2020meta}, recent works on augmentation of meta-training tasks have shown a marked improvement~\cite{ni2021data,yao2021improving,yao2021meta}.
The objective of task augmentation is to draw the empirical task distribution which is formed by assembling Dirac delta functions located in each meta-training task closer to the true task distribution. 
Consequently, achieving this objective requires a qualified task augmentation approach to simultaneously possess the following three properties: 
%%%%%%%%%%%%%%YING: make sure your wordings are consistent, task-awareness is a noun, while task-imaginary and model-adaptive are both adjectives. 
(1) %\textit{task-awareness}
\emph{task-aware}:
%(Figure~\ref{fig:1b}), which enables 
the augmented tasks %to distribute around the sampled tasks, making the sampling distribution a little closer to 
comply with
the true task distribution, being not erroneous to lead the meta-knowledge astray (tasks A and B in Figure~\ref{fig:intro}a);
%, i.e.,so that they are not erroneous to mislead the meta-knowledge} and also preventing the generation of erroneous tasks (i.e., out-of-distribution tasks);
(2) \textit{task-imaginary}:
the augmented tasks cover a substantial portion of the true distribution, embracing task diversity which  
task-awareness is nonetheless inadequate to guarantee (tasks C, D, and E in Figure~\ref{fig:intro}b); %(Figure~\ref{fig:1c}), which could generate new classes for the training tasks, making the training tasks more diverse and sparsely distributed over the task distribution, so as to minimize the learner overfitting; 
(3) \textit{model-adaptive}: the augmented tasks are timely in improving the current meta-knowledge, to which the meta-knowledge before augmentation struggles to generalize (task F in Figure~\ref{fig:intro}c). 
Unfortunately, developing such a qualified task augmentation approach remains challenging. First, the task-aware methods sacrifice task diversity for %establishing
task-awareness 
-- they establish task-awareness by injection of
%The existing augmentation-based methods usually adopt rotation or mixup operations to generate more tasks. Some methods belong to the task-aware methods. MetaAug~\cite{rajendran2020meta} imposes 
the same random noise to %training
labels of the support %set 
and query set~\cite{rajendran2020meta}, rotation of both support and query images~\cite{ni2021data}, or mix-up of
%in order to make the model more rely on the support set while predicting the query set.
%MetaMix~\cite{yao2021improving} augments a meta-training task by mixing 
support and query examples within %a
each
task~\cite{yao2021improving}, all of which result in augmented tasks that are within the immediate vicinity of sampled meta-training tasks as shown in  Figure~\ref{fig:intro}a. 
%However, since these methods do not generate a new semantic class of the support and the query, they could only generate limited imaginary tasks beyond sampled meta-training tasks.
Second, the task-imaginary method~\cite{yao2021meta} that mixes up both support examples of two distinct tasks and their query examples in the feature space compromises on task-awareness~-- the resulting examples are even multi-modal in Figure~\ref{fig:intro}e and constitute an erroneous task that fails to comply with the task distribution.
%Different from these task-aware augmentation methods, other methods are task-imaginary augmentation methods, which generate new tasks to alleviate the learner overfitting. Meta-Maxup~\cite{ni2021data} rotates both the support set and query set and views each rotation as a new task.\textcolor{red}{(if put this method to task-awareness?)}
%MLTI~\cite{yao2021meta}  interpolates the features of different tasks linearly to generate a new task. However, these methods generate more imaginary tasks at the expense of generating potential erroneous tasks, they cannot guarantee the validity of augmented tasks (Figure~\ref{fig:1c}). 
Third, how to adaptively augment tasks that maximally improve the meta-knowledge and thereby the performance on meta-testing tasks remains unexplored.
%Moreover, all current augmentation-based methods do not consider \textit{model-adaptive},
%%%%%%%%%%%%%%%%%%%%%YING: model-adaptive is an adjective, which can not be used as the direct object of an verb.
%which greatly limits the performance gains that task augmentation can bring.

%%%%%%%%%%%%%%%YING: why "aims"?? this is a grammatical error.
To this end,
%Aims on these problems, 
%%%%%%%%%%%%%%%YING: to better introduce your method in the introduction, you have to fully explicate the "why" and "how" part, e.g., why do you construct a large local task pool? (say, to cover a broad range of local task distribution) how do you construct the task pool?
we propose the Adversarial Task Up-sampling (ATU) framework to augment tasks that are aware of the task distribution, imaginary, and adaptive to the current meta-knowledge. Grounded on gradient-based meta-learning algorithms that are generally applicable to either regression or classification problems, ATU takes the initialization of the base learner as the meta-knowledge.
Concretely, ATU consists of a task up-sampling network whose input is a task itself and outputs are augmented tasks. To ensure that the augmented tasks are imaginary and meanwhile faithful to the underlying task distribution, we train the up-sampling network to minimize the Earth Mover Distance between augmented tasks and the local task distribution characterized by
%task augmentation framework capable of generating tasks that meet all 
%%%%%%%%%%%%%%%YING: again, all the words connected by ``and'' should be in the same POS; task-awareness is noun while the other two are adjectives.
%task-awareness, task-imaginary and model-adaptive needs. 
%For the proposed framework, motivated by the point-clouds up-sampling methods~\cite{yuan2018pcn}, we firstly construct a large local task pool (e.g., 
a set of sampled tasks.
%from task distribution) to represent the local task manifold. Then, we down-sample it to build a sparse task manifold as the input of the augmentation model,
%%%%%%%%%%%%%%%YING: "hoping" is very non-formal; avoid using it.
%hoping it could generate new tasks on the local task manifold under the guidance of distribution similarity. 
Besides, we enforce the up-sampling network to produce challenging tasks that complement the current initialization, by maximizing the loss of the model adapted from the initialization on their query examples and minimizing the similarity between the gradient of the initialization with respect to their support examples and that of their query examples.

%introduce the adversarial training on the augmentation network by exploiting the guidance of the meta model, hoping it could generate more hard or informative tasks on the local task manifold so as to maximize the benefits of task augmentation.  

% Aims on these problems, we propose a task augmentation framework capable of generating tasks that meet all task-awareness, task-imaginary and model-adaptive needs. For the proposed framework, motivated by the point-clouds up-sampling methods~\cite{yuan2018pcn}, we firstly construct a large local task pool (e.g., a set of sampled tasks from task distribution) to represent the local task manifold, and sampling a smaller task pool from it as the sparse task manifold. Then, we use the augmentation network to densify these sparse tasks and train the network with the earth mover’s distance (EMD) between the generated tasks and the local task manifold, so as to generate imaginary tasks. Besides, we exploit the adversarial training on the augmentation network to maximize the benefits by task augmentation. 
% three-fold
In summary, our main contributions are: (1)
we present the first task-level augmentation network that learns to
%propose a task augmentation framework that can 
generate tasks that %not only 
simultaneously
meet the %requirements 
qualifications
of being task-aware, task-imaginary, and model-adaptive; (2) %task awareness and task imagination, but also maximize the performance of the meta-model; 
%2) Theoretically, 
we provide a theoretical analysis to justify that the proposed ATU framework indeed promotes
%prove the reason why our augmentation methods are
task-awareness; (3) 
we conduct comprehensive
%regression and classification
experiments covering both regression and classification problems and a total of five datasets, where the proposed ATU improves 
%validate the superiority of the proposed method for improving 
the generalization ability of gradient-based meta-learning algorithms by up to 3.81\%.

%\textcolor{green}{This writing part done. Need to update the 3th paragraph after uploading the new figs}

% \ying{
% AdversariaL TAsk Imagination (ALTAI)\\
% Task-awareness: correct - minimize memorization\\
% Task-imagination: imaginary - minimize learner overfitting\\
% Model-adaptive: effective - maximize the benefits by task augmentation
% }
\section{Related Work}
\vspace{-6pt}
\begin{wraptable}[6]{r}{8.5cm}
\vspace{-24pt}
% \vspace{9pt}
\centering
\caption{\small Summary of existing task augmentation strategies.}
\label{tab:related_work}
    \resizebox{74mm}{11.2mm}{
 	    \begin{tabular}{c|c|c|c} 
			\toprule
			\multicolumn{1}{l}{\multirow{2}{*}{\textbf{Method}}} &
			\multicolumn{1}{c}{\multirow{2}{*}{\textbf{Task-aware}}} & 
			\multicolumn{1}{c}{\multirow{2}{*}{\textbf{Task-imaginary}}} & 
			\multicolumn{1}{c}{\multirow{2}{*}{\textbf{Model-adaptive}}}  \\
			\multicolumn{1}{l}{\multirow{2}{*}{}} &
			\multicolumn{1}{c}{\multirow{2}{*}{}} & 
			\multicolumn{1}{l}{\multirow{2}{*}{}} &
			\multicolumn{1}{c}{\multirow{2}{*}{}} \\
			\midrule
			\multicolumn{1}{l}{\multirow{1}{*}{MetaAug~\cite{rajendran2020meta}}} &
			\multicolumn{1}{c}{\multirow{1}{*}{\ding{51}}} & 
			\multicolumn{1}{c}{\multirow{1}{*}{\ding{55}}} &
			\multicolumn{1}{c}{\multirow{1}{*}{\ding{55}}} \\
			
			\multicolumn{1}{l}{\multirow{1}{*}{MetaMix~\cite{yao2021improving}}} &
			\multicolumn{1}{c}{\multirow{1}{*}{\ding{51}}} & 
			\multicolumn{1}{c}{\multirow{1}{*}{\ding{55}}} &
			\multicolumn{1}{c}{\multirow{1}{*}{\ding{55}}} \\

			\multicolumn{1}{l}{\multirow{1}{*}{Meta-Maxup~\cite{ni2021data}}} &
			\multicolumn{1}{c}{\multirow{1}{*}{\ding{51}}} & 
			\multicolumn{1}{c}{\multirow{1}{*}{\ding{55}}} &
			\multicolumn{1}{c}{\multirow{1}{*}{\ding{55}}} \\
            
			\multicolumn{1}{l}{\multirow{1}{*}{MLTI~\cite{yao2021meta}}} &
			\multicolumn{1}{c}{\multirow{1}{*}{\ding{55}}} & 
			\multicolumn{1}{c}{\multirow{1}{*}{\ding{51}}} &
			\multicolumn{1}{c}{\multirow{1}{*}{\ding{51}}} \\

			\multicolumn{1}{l}{\multirow{1}{*}{ATU}} &
			\multicolumn{1}{c}{\multirow{1}{*}{\ding{51}}} & 
			\multicolumn{1}{c}{\multirow{1}{*}{\ding{51}}} &
			\multicolumn{1}{c}{\multirow{1}{*}{\ding{51}}} \\			
			\bottomrule
		\end{tabular}
		}
% \end{table}
\end{wraptable}
As a paradigm that effectively adapts the meta-knowledge learned from past tasks to accelerate the learning of newones, meta-learning has sparked considerable interest, especially for few-shot learning. It  falls into four major strands based on what the meta-knowledge is, i.e., %:
optimizer-based methods~\cite{andrychowicz2016learning,wichrowska2017learned}, feed-forward methods~\cite{ravi2016optimization,ha2016hypernetworks,xu2020metafun}, metric-based %meta-learning
methods~\cite{snell2017prototypical,sung2018learning,vinyals2016matching} and gradient-based methods~\cite{finn2017model,li2017meta,yao2019hierarchically,lee2020learning},where the inner optimizer, the mapping function from the support set to the task-specific model, the
distance metric measuring the similarity between samples, and the parameter initialization are formulated as the meta-knowledge that enables quick adaptation to a task
within a small number of steps.  Our method is primarily evaluated  on gradient-based methods which enjoy wide adoption and applicability in either classification or regression problems.

\textbf{Within-task Overfitting.} %in Meta Learning.}
%Different from traditional supervised learning, there exist two overfitting problems in meta-learning, which are \textbf{within-task overfitting} and \textbf{meta-overfitting}~\cite{rajendran2020meta,yin2020meta}. 
Few-shot learning puts meta-learning, especially gradient-based methods which require optimization of high-dimensional parameters within each task, at risk of within-task overfitting. Some works %utilize regularizers to
tackle the %former 
problem %in several ways: by 
with various ways of reducing the number of parameters to adapt in the inner loop:
only updating the head~\cite{raghu2019rapid} or the feature extractor
%in the inner loops during training and testing, respectively
\cite{oh2020boil},
%%% [Ying's comment: put your citation close to the phrase that exactly describes the work]
%through limiting the uptable parameters,
learning data-dependent latent generative embeddings of parameters~\cite{rusu2018meta} or context parameters~\cite{zintgraf2019fast} to adapt, imposing gradient dropout~\cite{tseng2020regularizing}, and
%5compressing the task embeddings, combining dropout strategy or 
generating stochastic
input-dependent perturbations~\cite{ lee2019meta}. 
The other bunch of works alleviates the problem through data augmentation within each task. %s. 
Ni et al. 
\cite{ni2021data} %augmented
applies standard augmentation like Random Crop and CutMix onto
support %set and query set 
samples.
%through four mode data augmentation
%Sun et al.~\cite{sunmeda}
%%% [Ying's comment: you can use \citeauthor]
Sun et al.~\cite{sun2019meta}
and Zhang et al.~\cite{zhang2018metagan} proposed to
%generated 
generate
more data within a class via a %proposed 
ball generator and generative adversarial networks, respectively. 
%%% [Ying's comment: acronym like GAN should be used as its full name, when it first appears in your paper.]
%Zhang et al.~\cite{zhang2018metagan} %and Park et al.~\cite{park2020meta} augmented data through GAN and transferred variance information, respectively.
%%% [Ying's comment: for each paragraph of related works, you should summarize your difference to the works and well position yours.]
These techniques designed for within-task overfitting, however, have been proved to lend little support to meta-overfitting which we focus on.

\textbf{Meta-overfitting.}
Distinguished from traditional overfitting within a task, two types of meta-overfitting including memorization and learner overfitting have been pinpointed in ~\cite{yin2020meta,rajendran2020meta}.
Despite meta-regularization techniques~\cite{jamal2019task,yin2020meta} that limit the capacity of the meta-learner,
%For the latter one, to avoid overfitting on training tasks, Yin et al.~\cite{yin2020meta} proposed a meta-regularization objective based on information theory and Jamal et al.~\cite{jamal2019task} imposed an entropy regularization on meta-learner to prevent biased learning. Other methods proposed 
task augmentation strategies~\cite{rajendran2020meta,murty2021dreca,ni2021data,liu2020task,yao2021meta,yao2021improving} %,lee2021improving}. 
have emerged as more effective solutions to meta-overfitting. Table~\ref{tab:related_work} presents a summary of these strategies, except the strategies of large rotation~\cite{liu2020task} being part of Meta-Maxup~\cite{ni2021data} and DReCa~\cite{murty2021dreca} applicable to natural language inference tasks only.
MetaAug~\cite{rajendran2020meta} augments a task %Rajendran et al.~\cite{rajendran2020meta} constructed 
%augmented tasks 
%%% [Ying's comment: pay attention to articles.]
by adding a random noise on %the 
labels %space. 
of both support and query sets, and
MetaMix~\cite{yao2021improving}
%proposes mixing 
mixes support and query %features
examples
within a task. 
Such within-task augmentation guarantees the validity of augmented tasks, i.e., being task-aware, though it almost does not alter the mapping from the support to the query set, i.e., generating limited imaginary tasks beyond meta-training tasks.
Meta-Maxup~\cite{ni2021data} and MLTI~\cite{yao2021meta} approach this problem via the cross-task mixup method, unfavorably at the expense of erroneous tasks. Our work seeks a novel task augmentation framework capable of generating tasks that not only meet the task-awareness and task-imagination needs but also adapt to maximally benefit the up-to-the-minute meta-learner.

\section{Preliminaries}
\vspace{-1mm}
\subsection{Meta-Learning Problem and Gradient-Based Meta-Learning}
\vspace{-1mm}
Meta-Learning model $f$ are trained and evaluated on episodes of few-shot learning tasks. Assume the task distribution is $p(\mathcal{T})$. A few-shot learning task $T_i$ i.i.d. sampled from $p(\mathcal{T})$ consists of a support set $D_{i}^{s}=(X_i^{s},Y_i^{s})=\{(x_{i,j}^{s},y_{i,j}^{s})\}_{j=1}^{K^s}$
and a query set $D_{i}^{q}=({X_i^{q},Y_i^{q})}=\{(x_{i,j}^{q},y_{i,j}^{q})\}_{j=1}^{K^q}$, where $X_i^{s}$ and $Y_i^{s}$, ($X_i^{q}$ and $Y_i^{q}$) are the collection of inputs and labels in support (query) set, and $K^s$ ($K^q$) is the size of support (query) set.

%\kai{
The most representative gradient-based meta-learning algorithm is MAML~\cite{finn2017model}. MAML aims to learn an initialization parameter $\theta_0$ of the model $f$ that can be adapted to any new task after a few steps of gradient update. Concretely, given a specific task ${D_{i}^{s}, D_{i}^{q}}$ and a parametric model $f_{\theta}$, MAML initializes the model parameter $\theta$ by $\theta_0$ and updates $\theta$ by performing gradient descent on the support set $D_{i}^{s}$. It then optimizes the initialization parameter $\theta_0$ by minimizing the loss $\mathcal{L}$ estimated on the query set $D_{i}^{q}$. The objective of MAML can be formulated as
% \begin{equation}\label{true_risk}
%     \min_{\theta_0} \mathbb{E}_{\mathcal{T}_i\sim p(\mathcal{T})}\ \mathcal{L}(f_{\phi_i}(X_i^{q}),Y_i^{q}), 
%     \quad \quad \quad \text{s.t.} \quad \phi_i = \theta_0 - \alpha\nabla_{\theta_0} \mathcal{L}(f_{\theta_0}(X_i^{s}),Y_i^{s}).
% \end{equation}
\begin{equation}\label{true_risk}
    \min_{\theta_0} \mathbb{E}_{T_i\sim p(\mathcal{T})}\ \mathcal{L}(\phi_i, D_i^q), 
    \quad \quad \quad \text{s.t.} \quad \phi_i = \theta_0 - \alpha\nabla_{\theta_0} \mathcal{L}(\theta_0, D_i^s).
\end{equation}
\subsection{Earth Mover's Distance}
\vspace{-2mm}
To estimate the distance between two tasks, we use Earth Mover Distance (EMD). 
Earth Mover Distance, a.k.a. Wasserstein metric, is a distance measure of two probability distributions or two sets of points, and is widely used in image retrieval and point cloud up-sampling works~\cite{yuan2018pcn,yu2018pu}. 
Given two sets $S_1$ and $S_2$ with the same size, EMD calculates their distances as:
\begin{equation}\label{EMD}
 d_{EMD}(S_1,S_2) = \min_{\phi: S_1\rightarrow S_2} \frac{1}{\|S_1\|} \sum_{x\in S_1} \| x-\phi(x)\|_2,
\end{equation}
where $\phi$ is a bijective projection mapping $S_1$ to $S_2$. The value of EMD in Eq.\eqref{EMD} can be obtained by solving the linear programming problem w.r.t. $\phi$. %Other measures, such as Total Variation distance, Kullback-Leibler divergence, Jensen-Shannon divergence, could be used to estimate the distance between two sets. But compared to these method, EMD 

\section{Adversarial Task Up-sampling}
\vspace{-2mm}
In practice, the task distribution $p(\mathcal{T})$ is unknown and we optimize the meta parameter $\theta_0$ with an empirical estimation of Eq.\eqref{true_risk} over of meta-training tasks $\{T_{i}\}_{i=1}^{N_T}$ as
\begin{equation}\label{emp_risk}
    \min_{\theta_0} \frac{1}{N_T}\sum_{i=1}^{N_T} \mathcal{L}(\phi_i, D_i^q)), 
    \quad \quad \quad \text{s.t.} \quad \phi_i = \theta_0 - \alpha\nabla_{\theta_0} \mathcal{L}(\theta_0, D_i^s).
\end{equation}
Given a finite set of meta-training tasks, the empirical task distribution may deviate from the true task distribution. The meta-model trained on such a finite set of tasks will cause memorization or learner overfitting~\cite{yin2020meta,rajendran2020meta}, which hurts the generalization to new tasks. 
To alleviate this problem, we propose a new task up-sampling network to generate a sufficient number of diverse tasks such that the empirical task distribution formed by the original meta-training tasks and the augmented tasks together is closer to the true task distribution. To achieve this, the tasks generated by the task up-sampling network should match the true task distribution and cover a large fraction of it. %That is, the task augmentation network should be task-aware, task-imaginary. Moreover, it should also be model-adaptive to generate the critical tasks that improve the current meta model. 
However, since the true task distribution and its underlying manifold are unknown, we cannot provide the task up-sampling network with explicit information about it.
%we cannot explicitly characterize the task distribution and, therefore, cannot generate new tasks directly from the characterized task distribution. 
Instead, we generate new tasks by performing Task Up-sampling (TU) from a set of training tasks that implicitly comprise the latent task manifold information. The idea of Task Up-sampling is inspired by point cloud up-sampling methods~\cite{yuan2018pcn,yu2018pu}, which generate up-sampled points lying on the latent distribution (i.e., the shape) of the given local point patch.
%is proposed to generate up-sampled points that stay on the underlying surface of a given local point patch.
Similar to the point cloud up-sampling algorithm, our augmentation network receives a task patch consisting of a set of tasks $\mathcal{T}_p = \{T_i\}_{i=1}^{N_p}$ where $N_p$ is the set size, and generates up-sampled tasks $\mathcal{T}_{up}$ that are uniformly distributed over the same underlying task distribution as the task patch. 

% After receiving the task patch $\mathcal{T}_p$, we first extract the set feature $h_s$ of the input task sets by the set encoder $g_s$ as $h_s = g_s(\mathcal{T}_p)$. The set feature $h_s$ extracts the global information of the whole input task patch and is used to guide the task generation from a local task in the decoder. 
% We also use a set generator, the coarse generator $g_c$, to generates a set of coarse tasks $\mathcal{T}_c = g_c(\mathcal{T}_p)$ aiming to reconstruct the task distribution of the input task patch uniformly. 
% % When the tasks in a task patch are sufficiently different from each other, the coarse generator could be set as a identity mapping. Note that the coarse task set generates the same number of tasks as the input task patch.
% For each task $T_i^c$ in $\mathcal{T}_c$, the decoder $g_d$ generates $r$ tasks located around $T_i^c$ in the task manifold by taking as input $r$ random perturbances $\{z_i\}_{i=1}^r$ that are i.i.d. noise sampled from a uniform distribution and the set feature $h_s$ as input. 
\begin{figure}[t]
\centering
\includegraphics[width=5.45in, height = 1.3in]{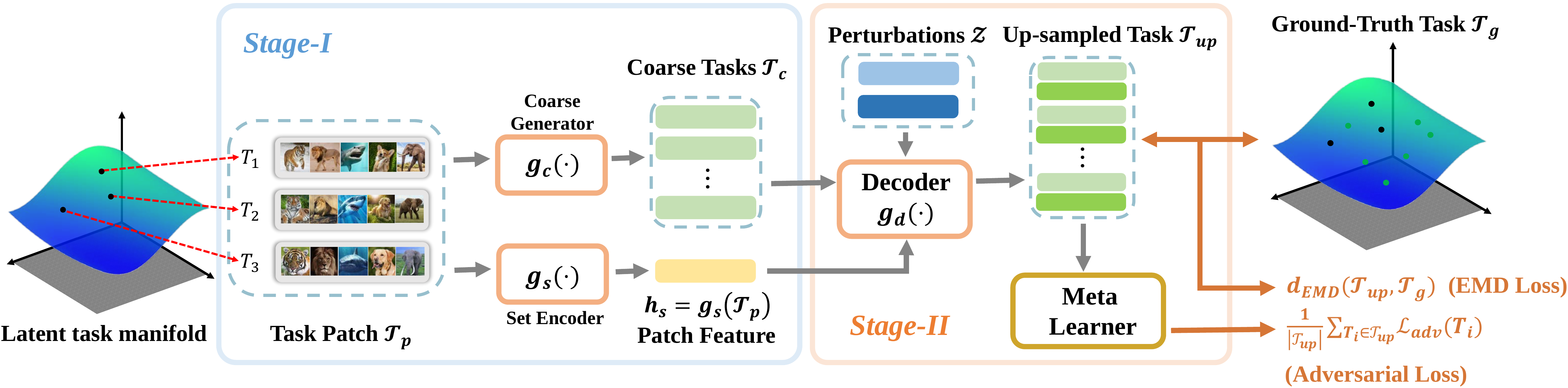}
\caption{Illustration of the ATU algorithm with a task up-sampling network. The task up-sampling network consists of a set encoder $g_s(\cdot)$ which extracts a set feature of the input task patch, a coarse task generator $g_c(\cdot)$ which generates coarse tasks given the task patch, and a decoder $g_d(\cdot)$ which generates fine tasks from coarse tasks based on random perturbance and the set feature.}
% \caption{Illustration of the Adversarial Task Up-sampling algorithms with a task augmentation network. The task augmentation network consists of a set encoder $g_s$ which extracts a set feature of the input task patch, a coarse task generator $g_c$ which generates coarse tasks given the task patch, and a decoder $g_d$ which generates fine tasks from coarse tasks based on random perturbance and the set feature.}
\label{framework}
\end{figure}
Due to the high complexity of task generation, it is infeasible to directly generate up-sampling tasks without sacrificing the quality of the tasks. Inspired by \cite{yuan2018pcn}, we propose a two-stage generation strategy to generate the up-sampled tasks. In the first stage, we produce a sparse set of tasks, aiming at recovering the \emph{global} task distribution of the task patch. The tasks obtained in the first stage are called coarse tasks. In the second stage, we generate multiple tasks for each coarse task, aiming to characterize the \emph{local} task distribution around each coarse task. To guide the second generation stage, we use the patch features of the input task set as input to provide global task information and also multiple random noise vectors as input to provide directional perturbations to generate diverse tasks around the coarse task. 
The generation process is summarized in Fig.~\ref{framework}.

Our proposed task up-sampling network consists of 3 components, namely, a coarse task generator $g_c(\cdot)$, a set encoder $g_s(\cdot)$, and a decoder $g_d(\cdot)$. 
The coarse task generator is similar to a set auto-encoder. It first encodes the information of the whole input task set and then decodes it to generate $r_c N_p$ coarse tasks $\mathcal{T}_c = \{T^c_i\}$. %Generally, the number of coarse tasks is the same as the number of input tasks, or twice the number of input tasks the number of input tasks. 
The set encoder, denoted by $g_s(\cdot)$, extracts the set information of the input task patch as a patch feature $h_s$ to provide global information in the second-stage generation. 
For each task $T_i^c$ in $\mathcal{T}_c$, the decoder $g_d(\cdot)$ generates $r_d$ tasks located around $T_i^c$ in the task manifold by taking as input $r_d$ random perturbations $\{z_i\}_{i=1}^r$ that are i.i.d. noise sampled from a uniform distribution and the set feature $h_s$ as input.  
% The perturbation $z_i$ provides the randomness of generating different tasks and the set feature $h_s$ provides the global information about the entire task patch.
In general, we use the same perturbations for each coarse task $T_i^c$ in $\mathcal{T}_c$. Finally, we obtain the up-sampled task set $\mathcal{T}_{up}$ consisting of $r N_p$ tasks as $\mathcal{T}_{up} = g_d(\mathcal{T}_c, \mathcal{Z}, h_s)$, where $r=r_c \times r_d$ is the up-sampling ratio.%, where $\mathcal{Z} = \{z_i\}_{i=1}^r$ is the perturbation set. 
 We denote the task augmentation network by $G_{\theta_g}(\mathcal{T}_p, \mathcal{Z})$ where $\theta_g$ is the trainable parameters of the task augmentation network.

% \begin{figure}[t]
% \centering
% \includegraphics[width=5.45in, height = 1.3in]{./Figs/method/Task_Aug1.pdf}
% % \includegraphics[width=5.35in, height = 1.45in]{./Figs/method/Task_Aug1.pdf}
% \caption{Illustration of the ATU algorithm with a task up-sampling network. The task up-sampling network consists of a set encoder $g_s$ which extracts a set feature of the input task patch, a coarse task generator $g_c$ which generates coarse tasks given the task patch, and a decoder $g_d$ which generates fine tasks from coarse tasks based on random perturbance and the set feature.}
% % \caption{Illustration of the Adversarial Task Up-sampling algorithms with a task augmentation network. The task augmentation network consists of a set encoder $g_s$ which extracts a set feature of the input task patch, a coarse task generator $g_c$ which generates coarse tasks given the task patch, and a decoder $g_d$ which generates fine tasks from coarse tasks based on random perturbance and the set feature.}
% \label{framework}
% \end{figure}

%\kai{In each iteration of the training phase, we randomly extract $r N_p$ tasks from the meta-training task set to form the ground truth task set $\mathcal{T}_g$.}

In each iteration of the training phase, we construct $r N_p$ tasks to form the ground truth tasks set $\mathcal{T}_g$ (e.g., randomly select from the meta training task set).      Then we sample $N_p$ tasks from $\mathcal{T}_g$ to form the task patch $\mathcal{T}_p$ and randomly sample $r$ perturbation noise vectors to form the perturbation set $\mathcal{Z} = \{z_i\}_{i=1}^r$. By feeding 
$\mathcal{T}_p$ and $\mathcal{Z}$ to the task augmentation network, we obtain the up-sampled task set $\mathcal{T}_{up} = G_{\theta_g}(\mathcal{T}_p, \mathcal{Z})$. 
To train the task augmentation network, we apply EMD loss between the up-sampled task set $\mathcal{T}_{up}$ and the ground-truth task set $\mathcal{T}_g$ to encourage the generated task set to have the same distribution as the true task distribution. However, it may still be insufficient to make the up-sampled tasks cover a significant fraction of the true task distribution. In this case, the up-sampling tasks provide limited additional information compared with the original meta-training tasks, and thus the meta-learner has restricted benefit from the generated tasks.
%recover the empirical task distribution formed by the $N_T$ training tasks. In this case, nearly no more information is provided in the up-sampled tasks than in the original meta-training tasks, and the meta-learner has limited benefit from the generated tasks. 
To generate more informative tasks for the meta-learner, we want the generated tasks to be difficult for the current meta model $\theta_0$. Following \cite{yao2021meta1}, we measure the difficulty of the a task for $\theta_0$ by the loss estimated on query set w.r.t. to $\phi_i$, i.e. $\mathcal{L}(\phi_i, D_i^q)$, and the gradient similarity between the support and query sets w.r.t. $\theta_0$, i.e. $\langle \nabla_{\theta_0} \mathcal{L}(\theta_0, D_i^s), \nabla_{\theta_0} \mathcal{L}(\theta_0, D_i^q)\rangle$. The large loss and small gradient similarity indicate a difficult task. Therefore, we want to maximize the following objective function to generate informative tasks:
\begin{equation}\label{eq:4}
	\mathcal{L}_{adv}(\theta_0, (D_i^s, D_i^q)) = \eta_1 \mathcal{L}(\phi_i, D_i^q) - \eta_2 \langle \nabla_{\theta_0} \mathcal{L}(\theta_0, D_i^s), \nabla_{\theta_0} \mathcal{L}(\theta_0, D_i^q)\rangle,
\end{equation}
where $\eta_1$ and $\eta_2$ are two hyperparameters that control the strength of the two terms in $\mathcal{L}_{adv}$. We call this loss adversarial loss because it aims to increase the difficulty of the up-sampling tasks for the meta-learner while the meta-learner is trained to minimize the loss on the generated difficult tasks. And we named the proposed algorithm as Adversarial Task Up-sampling (ATU).

Together with the EMD loss, we obtain the objective to train the task up-sampling network:
\begin{equation}\label{Trainingloss}
	\mathcal{L}_{ATU}(\theta_g, \mathcal{T}_p) =  d_{EMD}(\mathcal{T}_{up}, \mathcal{T}_g) - \frac{1}{rN_p} \sum_{T_i \in \mathcal{T}_{up}} \mathcal{L}_{adv}(\theta_0, (D_i^s, D_i^q)).
\end{equation}
Note that the gradient of Eq.~\eqref{Trainingloss} will not be backpropagated to the meta model $\theta_0$ and the meta model will be updated by minimizing the meta loss in Eq.~\eqref{emp_risk} on the up-samples tasks $\mathcal{T}_u$.
We summarize the proposed ATU in Algorithm 1 in Appendix A. %\textbf{XXXXXXXXXXXXXXXXXXXX}. 

%\kai{(better put in the experiment section.)}
\subsection{ ATU on Regression and Classification Problem}
\textbf{Regression Tasks.} We consider a simple regression problem: sinusoidal regression, which is widely used to evaluate the effectiveness of the meta-learning methods. In sinusoidal regression problem, before feeding a task $T_i = {(D_i^s, D_i^q)}$ to the task up-sampling network, we represent the task by a connection vector as $[x^s_1, y^s_1, x^s_2, y^s_2, ..., x^s_{K^s}, y^s_{K^s}, x^q_1, y^q_1, x^q_2, y^q_2, ..., x^q_{K^q}, y^q_{K^q}]$. In particular, we sort the support set and query set  by the input respectively, such that $x^s_1 \le x^s_2 \le ... \le x^s_{K^s}$ and $x^q_1 \le x^q_2 \le ... \le x^q_{K^q}$. After sorting, the task input is invariant to the permutation of data in support and query sets, and the feature extracted for each task is a permutation-invariant feature. This simplifies the design of the task up-sampling network. To generate the coarse tasks, we first use the set encoder $g_s$ to extract the set feature $h_s$ and directly generate the coarse tasks from the set feature $h_s$. For the regression problem, we add an extract EMD loss on the support and query set for each generated task $T_i^u \in \mathcal{T}_u$ to encourage the points in the generated support set and query set to belong to the same sinusoidal task, and the objective becomes:
\begin{equation}
	\mathcal{L}_{ATU}(\theta_g, \mathcal{T}_p) =  d_{EMD}(\mathcal{T}_{up}, \mathcal{T}_g) + \eta_3 \frac{1}{rN_p} \sum_{T_i \in \mathcal{T}_u} d_{EMD}(D_i^s, D_i^q) - \frac{1}{rN_p} \sum_{T_i \in \mathcal{T}_u} \mathcal{L}_{adv}(\theta_0, (D_i^s, D_i^q).
	\label{regression_augnetwork_loss}
\end{equation}
After generating the up-sampled tasks, we transform the task representation feature, which is a concatenating vector, back to a support set and a query set.

\textbf{Classification Tasks.} 
For an $N$-way $K^s$-shot classification problem with $K^q$ query samples for each class, we split the task into $(K^s\!\!+\!\!K^q)$ $N$-way $1$-shot classification tasks without query examples. We represent the task by concatenating the input from $N$ classes in a fixed order (based on classes). We treat the $(K^s\!\!+\!\!K^q)$ tasks as a task patch and feed them to the task up-sampling network. 
Since the task distribution of the image classification problem is extremely complex, it is impractical to generate the coarse tasks from a set feature. Instead, we use the original tasks as the coarse tasks and generate the up-sampled tasks by a more informative perturbance around the original tasks. To achieve this, we generate the perturbation by randomly sampling extra $K_M$ images from $N$ different classes in the base set. Then for the image $x_i$ in a class of a task in the coarse tasks $\mathcal{T}_c$, we subtract it from the $K_M$ image to obtain $K_M$ residual images, concatenate each residual image with a noise vector, and use an attention network to obtain a residual images feature $x_i^{res}$ for the image $x_i$. Finally, we generate the image as $x_i^u\!=\! x_i \!+\! x_i^{res}$ for the augmented task. We repeat this process for all images in a coarse task to get an augmented task and apply the $r$ noise vector to get $r$ up-sampled tasks. More details of the network structures and the training details are shown in Appendix B.

\label{sec:method}

\section{Theoretical Analysis}
We will introduce the formal definition of an up-sampled task that conforms to task-awareness, based on which we present the essential property of our proposed ATU framework in maximizing the task-awareness, compared to previous task augmentation approaches.
\begin{definition}[Task-aware Up-sampling]
Suppose that we are given a set of $N_p$ tasks $\{\mathbf{X}_i,\mathbf{Y}_i\}_{i=1}^{N_{p}}$ from which we up-sample a new task $T_{up}$. For each $i$-th task,  its ground-truth parameter that map the input $\mathbf{X}_i$ to the output $\mathbf{Y}_i$ is $\theta_{i}$, i.e., $\mathbf{Y}_i=f_{\theta_{i}}(\mathbf{X}_i)$.
The up-sampled task $T_{up}=\{\mathbf{X}_u,\mathbf{Y}_u\}$ is defined to be task-aware, if and only if $\theta_u=g(\theta_1,\cdots,\theta_{N_u})$ and   $\mathbf{Y}_u = f_{\theta_u}(\mathbf{X}_u)$ where $g$ is the up-sampling function and $\theta_u$ is the up-sampled parameter.
\end{definition}
This definition states two prerequisites a task-aware up-sampling has to meet: (1) the up-sampling is performed in the functional space, which is to relate $N_u$ parameters via $g$; (2) the mapping between the input and the output of an up-sampled task satisfies $f_{\theta_u}$.
\begin{property}[Task-awareness Maximization]
Consider $N_u=2$, $g(\theta_1,\theta_2)=(1-\lambda)\theta_1 + \lambda \theta_2$, $f_{\theta_1}(\cdot)=\mathbf{W}_1$, and  $f_{\theta_2}(\cdot)=\mathbf{W}_2$. The proposed ATU algorithm that pursues an up-sampled task $T_{up}=\{\mathbf{X}_u,\mathbf{Y}_u\}$ via minimizing the EMD loss between $T_1$ and $T_2$ maximizes the task-awareness, i.e., minimizing the distance between $\mathbf{Y}_u$ and $f_{\theta_u}(\mathbf{X}_u)$.
\end{property}

\begin{proof}
According to the definition of EMD(Eq.~(\ref{EMD})), it solves: $\phi^*\!=\!\arg\min_{\phi\in\boldsymbol{\Phi}}\sum_{j}\!\Vert \mathbf{x}_{1,j}\!-\!\mathbf{x}_{2,\phi(j)}\!\Vert_2$, where $\boldsymbol{\Phi}=\{\{1,\cdots,n_1\}\mapsto\{1,\cdots,n_2\}\}$ 
%$\boldsymbol{\Phi}=\{\{1,\cdots,n\}\mapsto\{1,\cdots,n\}\}$ 
denotes the set containing all possible bijective assignments, each of which gives one-to-one correspondence between $T_1$ and $T_2$.
Based on the optimal assignments $\phi^*$, the EMD is known to be defined as: % $d_{EMD}=\frac{1}{\min\{n_1, n_2\}}\sum_{j}\Vert \mathbf{x}_{1,j}-\mathbf{x}_{2,\phi^*(j)}\Vert_2$.
\begin{equation*}
    d_{EMD}=\frac{1}{\min\{n_1, n_2\}}\sum_{j}\Vert \mathbf{x}_{1,j}-\mathbf{x}_{2,\phi^*(j)}\Vert_2
\end{equation*}
Since we train the up-sampling network via minimizing the EMD loss $d_{EMD}$ between $\{T_u\}$ and $\{T_1, T_2\}$ to learn their local manifold, we reasonably conclude that one feasible solution for the up-sampled task $T_u$ follows:   
$\mathbf{y}_{u,j} = (1-\lambda)\mathbf{y}_{1,j} + \lambda\mathbf{y}_{2,\phi^*(j)},  \mathbf{x}_{u,j} = (1-\lambda)\mathbf{x}_{1,j} + \lambda\mathbf{x}_{2,\phi^*(j)}.$ 

With this feasible $\mathcal{T}_u$, we evaluate the task-awareness, the distance between $\mathbf{Y}_u$ and {\small $f_{\theta_u}(\mathbf{X}_u)$}, to be \\
 $~~~~~~~~~~~\Vert\mathbf{Y}_u-f_{\theta_u}(\mathbf{X}_u)\Vert_2  =\sum_{j}\Vert \mathbf{y}_{u,j} - f_{\theta_u}(\mathbf{x}_{u,j} )\Vert_2$ \\[2mm]
= $\sum_{j}\Vert(1-\lambda)\mathbf{y}_{1,j} + \lambda\mathbf{y}_{2,\phi^*(j)} $\nonumber -$[(1-\lambda)\mathbf{W}_1+\lambda\mathbf{W}_2][(1-\lambda)\mathbf{x}_{1,j} + \lambda\mathbf{x}_{2,\phi^*(j)}]\Vert_2$ \nonumber \\[1.4mm]
= $\sum_{j}\Vert(1-\lambda)\mathbf{W}_1\mathbf{x}_{1,j} + \lambda\mathbf{W}_2\mathbf{x}_{2,\phi^*(j)} $\nonumber -$[(1-\lambda)\mathbf{W}_1+\lambda\mathbf{W}_2][(1-\lambda)\mathbf{x}_{1,j} + \lambda\mathbf{x}_{2,\phi^*(j)}]\Vert_2$ \nonumber \\[1.4mm]
= $\sum_j\lambda^2(1-\lambda)^2\Vert(\mathbf{W_1}-\mathbf{W}_2)(\mathbf{x}_{1,j}-\mathbf{x}_{2,\phi^*(j)})\Vert_2$ %\nonumber \\[1.4mm]
$\leq \sum_j\lambda^2(1-\lambda)^2\Vert\mathbf{W_1}-\mathbf{W}_2\Vert_2\Vert\mathbf{x}_{1,j}-\mathbf{x}_{2,\phi^*(j)}\Vert_2 $ \nonumber\\
%\leq n^2 \lambda^2(1-\lambda)^2\Vert\mathbf{W_1}-\mathbf{W}_2\Vert_2 \ d_{EMD}.\nonumber\\
$\leq n_1 n_2 \lambda^2(1-\lambda)^2\Vert\mathbf{W_1}-\mathbf{W}_2\Vert_2 \ d_{EMD}.$\nonumber\\

Therefore, by minimizing the EMD loss $d_{EMD}$, we minimize the distance between $\mathbf{Y}_u$ and $f_{\theta_u}(\mathbf{X}_u)$; in other words, the task-awareness is maximized.
\end{proof}
\vspace{-3mm}

    % \begin{multiline}
    % \begin{equation*}
    % \small \qquad  $\Vert\mathbf{Y}_u-f_{\theta_u}(\mathbf{X}_u)\Vert_2  =\sum_{j}\Vert \mathbf{y}_{u,j} - f_{\theta_u}(\mathbf{x}_{u,j} )\Vert_2$ \\[2mm]
    %     = $\sum_{j}\Vert(1-\lambda)\mathbf{y}_{1,j} + \lambda\mathbf{y}_{2,\phi^*(j)} $\nonumber -$[(1-\lambda)\mathbf{W}_1+\lambda\mathbf{W}_2][(1-\lambda)\mathbf{x}_{1,j} + \lambda\mathbf{x}_{2,\phi^*(j)}]\Vert_2$ \nonumber \\[1.4mm]
    %     = $\sum_{j}\Vert(1-\lambda)\mathbf{W}_1\mathbf{x}_{1,j} + \lambda\mathbf{W}_2\mathbf{x}_{2,\phi^*(j)} $\nonumber -$[(1-\lambda)\mathbf{W}_1+\lambda\mathbf{W}_2][(1-\lambda)\mathbf{x}_{1,j} + \lambda\mathbf{x}_{2,\phi^*(j)}]\Vert_2$ \nonumber \\[1.4mm]
    %     = $\sum_j\lambda^2(1-\lambda)^2\Vert(\mathbf{W_1}-\mathbf{W}_2)(\mathbf{x}_{1,j}-\mathbf{x}_{2,\phi^*(j)})\Vert_2$ \nonumber \\[1.4mm]
    %     $\leq \sum_j\lambda^2(1-\lambda)^2\Vert\mathbf{W_1}-\mathbf{W}_2\Vert_2\Vert\mathbf{x}_{1,j}-\mathbf{x}_{2,\phi^*(j)}\Vert_2 $ \nonumber\\
    %     %\leq n^2 \lambda^2(1-\lambda)^2\Vert\mathbf{W_1}-\mathbf{W}_2\Vert_2 \ d_{EMD}.\nonumber\\
    %     $\leq n_1 n_2 \lambda^2(1-\lambda)^2\Vert\mathbf{W_1}-\mathbf{W}_2\Vert_2 \ d_{EMD}.$\nonumber\\
    % \end{equation*}
    % \end{multiline}
Previous task augmentation approaches directly mix up two tasks without minimizing the EMD loss, i.e., $ \mathbf{y}_{u,j} = (1-\lambda)\mathbf{y}_{1,j} + \lambda\mathbf{y}_{2,j}, 
    \mathbf{x}_{u,j} = (1-\lambda)\mathbf{x}_{1,j} + \lambda\mathbf{x}_{2,j}$.
In this case, the task-awareness is unwarranted as we have illustrated in Section~\ref{sec:introduction}, provided that $(\mathbf{W_1}-\mathbf{W}_2)(\mathbf{x}_{1,j}-\mathbf{x}_{2,j})=0$,  $\forall j$ seldom holds.

\label{sec:analysis}

\section{Experiments}
\vspace{-5pt}
To evaluate the effectiveness of ATU, we conduct extensive experiments to answer the following  questions: 
\textbf{Q1:} How does ATU perform compared to state-of-the-art task-augmentation-based and regularization meta-learning methods?
\textbf{Q2:} Whether can the proposed ATU consistently improve performance for different meta-learning methods? 
\textbf{Q3:} What does up-sampled task by ATU looks like?
\textbf{Q4:} What is the influence of increasing the task number within meta-training data on the performance improvement of ATU?  ~\textbf{Benchmarks.} We compared ATU with %current 
state-of-the-art task augmentation strategies for meta-learning, including MetaAug~\cite{rajendran2020meta}, MetaMix~\cite{yao2021improving}, Meta-Maxup~\cite{ni2021data}, MLTI~\cite{yao2021meta}, and regularization methods, including MetaDropout~\cite{lee2019meta}, TAML~\cite{jamal2019task}, and Meta-Reg~\cite{yin2020meta} for both regression and classification problems. We also consider a variant of ATU which removes the adversarial loss $\mathcal{L}_{adv}$ and trains the task augmentation network only through the EMD loss. We denote this variant by TU.
To validate the consistent effect of ATU in improving different meta-learners, we apply ATU and AU on MAML~\cite{li2017meta}, MetaSGD~\cite{li2017meta} and ANIL~\cite{raghu2019rapid}. We also consider cross-domain settings where the meta-testing tasks are from different domains.

\begin{minipage}[t]{\textwidth}
    \begin{minipage}[b]{0.45\textwidth}
        \centering
    \makeatletter\def\@captype{table}
    \caption{MSE with $\pm95\%$ confidence intervals on sinusoidal regression.}% different
    \label{Regression}
    \resizebox{65mm}{23mm}{
 	    \begin{tabular}{ccc|cc|cc|cc} 
			\toprule[1pt]
			\multicolumn{3}{l}{\multirow{1}{*}{Model% (Testing)
			}} & 
			\multicolumn{2}{c}{\multirow{1}{*}{10-shot}} &
			\multicolumn{2}{c}{\multirow{1}{*}{20-shot}} &
			\multicolumn{2}{c}{\multirow{1}{*}{30-shot}} 
 			 \\ \hline \hline
			\multicolumn{3}{l}{\multirow{1}{*}{DropGrad~\cite{tseng2020regularizing}}} &
            \multicolumn{2}{c}{\multirow{1}{*}{$0.91\pm0.17$}} &
			\multicolumn{2}{c}{\multirow{1}{*}{$0.62\pm0.12$}} &
			\multicolumn{2}{c}{\multirow{1}{*}{$0.55\pm0.13$}}  
		\\ 
			\multicolumn{3}{l}{\multirow{1}{*}{MetaAug~\cite{rajendran2020meta}}} &
            \multicolumn{2}{c}{\multirow{1}{*}{$0.93\pm0.18$}} &
			\multicolumn{2}{c}{\multirow{1}{*}{$0.65\pm0.14$}} & 
			\multicolumn{2}{c}{\multirow{1}{*}{$0.58\pm0.12$}}  
		\\ \hline
		    \multicolumn{1}{c}{\multirow{1}{*}{\qquad\quad\textit{Meta-Learner: MAML% as backbone
		    }}}   \\ 
		
			\multicolumn{3}{l}{\multirow{1}{*}{MAML~\cite{finn2017model}}} &
            \multicolumn{2}{c}{\multirow{1}{*}{$0.93\pm0.18$}} &
			\multicolumn{2}{c}{\multirow{1}{*}{$0.65\pm0.13$}} & 
			\multicolumn{2}{c}{\multirow{1}{*}{$0.58\pm0.12$}}  
		\\ 
			\multicolumn{3}{l}{\multirow{1}{*}{MetaMix~\cite{yao2021improving}}} &
            \multicolumn{2}{c}{\multirow{1}{*}{$0.81\pm0.17$}} &
			\multicolumn{2}{c}{\multirow{1}{*}{$0.58\pm0.12$}} & 
			\multicolumn{2}{c}{\multirow{1}{*}{$0.56\pm0.11$}}  
		\\ 	
			\multicolumn{3}{l}{\multirow{1}{*}{MLTI~\cite{yao2021meta}}} &
            \multicolumn{2}{c}{\multirow{1}{*}{$0.92\pm0.17$}} &
			\multicolumn{2}{c}{\multirow{1}{*}{$0.65\pm0.13$}} & 
			\multicolumn{2}{c}{\multirow{1}{*}{$0.62\pm0.12$}}  
		\\ 				
			\multicolumn{3}{l}{\multirow{1}{*}{TU}} &
            \multicolumn{2}{c}{\multirow{1}{*}{$0.84\pm0.16$}} &
			\multicolumn{2}{c}{\multirow{1}{*}{$0.55\pm0.12$}} & 
			\multicolumn{2}{c}{\multirow{1}{*}{$0.47\pm0.10$}}  
		\\ 						
			\multicolumn{3}{l}{\multirow{1}{*}{ATU}} &
            \multicolumn{2}{c}{\multirow{1}{*}{\textbf{0.70}~$\pm$~\textbf{0.14}}} &
			\multicolumn{2}{c}{\multirow{1}{*}{\textbf{0.47}~$\pm$~\textbf{0.13}}} & 
			\multicolumn{2}{c}{\multirow{1}{*}{\textbf{0.42}~$\pm$~\textbf{0.11}}}  
		\\ \hline
		    \multicolumn{1}{c}{\multirow{1}{*}{\qquad\quad\textit{Meta-Learner: MetaSGD% as backbone
		    }}}   \\ 
		
			\multicolumn{3}{l}{\multirow{1}{*}{MetaSGD~\cite{li2017meta}}} &
            \multicolumn{2}{c}{\multirow{1}{*}{$0.70\pm0.17$}} &
			\multicolumn{2}{c}{\multirow{1}{*}{$0.49\pm0.11$}} & 
			\multicolumn{2}{c}{\multirow{1}{*}{$0.42\pm0.09$}}  
		\\ 
			\multicolumn{3}{l}{\multirow{1}{*}{MetaMix~\cite{yao2021improving}}} &
            \multicolumn{2}{c}{\multirow{1}{*}{$0.60\pm0.15$}} &
			\multicolumn{2}{c}{\multirow{1}{*}{$0.37\pm0.09$}} & 
			\multicolumn{2}{c}{\multirow{1}{*}{$0.37\pm0.08$}}  
		\\ 	 
			\multicolumn{3}{l}{\multirow{1}{*}{MLTI~\cite{yao2021meta}}} &
            \multicolumn{2}{c}{\multirow{1}{*}{$0.66\pm0.16$}} &
			\multicolumn{2}{c}{\multirow{1}{*}{$0.51\pm0.11$}} & 
			\multicolumn{2}{c}{\multirow{1}{*}{$0.44\pm0.10$}} 
		\\ 				
			\multicolumn{3}{l}{\multirow{1}{*}{TU}} &
            \multicolumn{2}{c}{\multirow{1}{*}{$0.54\pm0.11$}} &
			\multicolumn{2}{c}{\multirow{1}{*}{$0.36\pm0.08$}} & 
			\multicolumn{2}{c}{\multirow{1}{*}{$0.31\pm0.08$}}  
		\\ 						
			\multicolumn{3}{l}{\multirow{1}{*}{ATU}} &
            \multicolumn{2}{c}{\multirow{1}{*}{$\textbf{0.49}$~$\pm$~$\textbf{0.10}$}} &
			\multicolumn{2}{c}{\multirow{1}{*}{$\textbf{0.34}$~$\pm$~$\textbf{0.08}$}} & 
			\multicolumn{2}{c}{\multirow{1}{*}{$\textbf{0.29}$~$\pm$~$\textbf{0.08}$}}  
        \\ 
			\bottomrule[1pt]
		\end{tabular}
		}  
        \end{minipage}
        \hspace{0.10in}
        \begin{minipage}[b]{0.45\textwidth}
        \centering
    \makeatletter\def\@captype{table}
        \caption{MSE with $\pm95\%$ confidence intervals on % different domain settings.
         cross-domain sinusoidal regression.}
        \label{Regression-cross} 
        
    \resizebox{67mm}{23mm}{
 	\begin{tabular}{ccc|ccc|ccc|ccc} 
		%\begin{tabular}{ccc|ccc|ccc} 
			\toprule[1pt]
			\multicolumn{3}{l}{\multirow{2}{*}{Cross-domain}} & 
			\multicolumn{3}{c}{\multirow{1}{*}{Frequency}} &
			\multicolumn{3}{c}{\multirow{1}{*}{Aplitude}} &
			\multicolumn{3}{c}{\multirow{1}{*}{Phase}} 
 			 \\ 
			\multicolumn{3}{l}{\multirow{2}{*}{}} & 
			\multicolumn{3}{c}{\multirow{1}{*}{[0.4,0.8]}} &
			\multicolumn{3}{c}{\multirow{1}{*}{[5.0,6.0]}} &
			\multicolumn{3}{c}{\multirow{1}{*}{[$-\pi$,0]}} 
 			 \\  			 
 			 \hline \hline
		    \multicolumn{1}{c}{\multirow{1}{*}{\qquad\quad\textit{Meta-Learner: MAML% as backbone
		    }}}   \\ 
		
			\multicolumn{3}{l}{\multirow{1}{*}{MAML~\cite{finn2017model}}} &
            \multicolumn{3}{c}{\multirow{1}{*}{$1.78\pm0.35$}} &
			\multicolumn{3}{c}{\multirow{1}{*}{$3.52\pm0.35$}} & 
			\multicolumn{3}{c}{\multirow{1}{*}{$3.12\pm0.52$}}  
		\\ 
			\multicolumn{3}{l}{\multirow{1}{*}{MetaMix~\cite{yao2021improving}}} &
            \multicolumn{3}{c}{\multirow{1}{*}{$1.67\pm0.30$}} &
			\multicolumn{3}{c}{\multirow{1}{*}{$3.60\pm0.28$}} & 
			\multicolumn{3}{c}{\multirow{1}{*}{$3.14\pm0.54$}}  
		\\ 	
		    % lam ~ B(2,2)
			\multicolumn{3}{l}{\multirow{1}{*}{MLTI~\cite{yao2021meta}}} &
            \multicolumn{3}{c}{\multirow{1}{*}{$1.92\pm0.42$}} &
			\multicolumn{3}{c}{\multirow{1}{*}{$3.56\pm0.37$}} & 
			\multicolumn{3}{c}{\multirow{1}{*}{$3.66\pm0.63$}}  
		\\ 				
			\multicolumn{3}{l}{\multirow{1}{*}{TU}} &
            \multicolumn{3}{c}{\multirow{1}{*}{$1.70\pm0.34$}} &
			\multicolumn{3}{c}{\multirow{1}{*}{$3.22\pm0.31$}} & 
			\multicolumn{3}{c}{\multirow{1}{*}{$2.88\pm0.48$}}  
		\\ 						
			\multicolumn{3}{l}{\multirow{1}{*}{ATU}} &
            \multicolumn{3}{c}{\multirow{1}{*}{\textbf{1.58}~$\pm$~\textbf{0.35}}} &
			\multicolumn{3}{c}{\multirow{1}{*}{\textbf{2.92}~$\pm$~\textbf{0.29}}} & 
			\multicolumn{3}{c}{\multirow{1}{*}{\textbf{2.58}~$\pm$~\textbf{0.48}}}  
		\\ \hline
		    \multicolumn{1}{c}{\multirow{1}{*}{\qquad\quad\textit{Meta-Learner: MetaSGD% as backbone
		    }}}   \\ 
		    
			\multicolumn{3}{l}{\multirow{1}{*}{MetaSGD~\cite{li2017meta}}} &
            \multicolumn{3}{c}{\multirow{1}{*}{$2.24\pm0.46$}} &
			\multicolumn{3}{c}{\multirow{1}{*}{$2.42\pm0.32$}} & 
			\multicolumn{3}{c}{\multirow{1}{*}{$2.73\pm0.56$}}  
		\\ 
			\multicolumn{3}{l}{\multirow{1}{*}{MetaMix~\cite{yao2021improving}}} &
            \multicolumn{3}{c}{\multirow{1}{*}{$1.77\pm0.35$}} &
			\multicolumn{3}{c}{\multirow{1}{*}{$2.50\pm0.27$}} & 
			\multicolumn{3}{c}{\multirow{1}{*}{$2.46\pm0.48$}}  
		\\ 	 

			\multicolumn{3}{l}{\multirow{1}{*}{MLTI~\cite{yao2021meta}}} &
            \multicolumn{3}{c}{\multirow{1}{*}{$1.80\pm0.42$}} &
			\multicolumn{3}{c}{\multirow{1}{*}{$2.56\pm0.28$}} & 
			\multicolumn{3}{c}{\multirow{1}{*}{$2.54\pm0.54$}} 
		\\ 				
			\multicolumn{3}{l}{\multirow{1}{*}{TU}} &
            \multicolumn{3}{c}{\multirow{1}{*}{$\textbf{1.64}$~$\pm$~$\textbf{0.38}$}} &
			\multicolumn{3}{c}{\multirow{1}{*}{$2.37\pm0.25$}} & 
			\multicolumn{3}{c}{\multirow{1}{*}{$\textbf{2.04}$~$\pm$~$\textbf{0.46}$}}  
		\\ 						
			\multicolumn{3}{l}{\multirow{1}{*}{ATU}} &
            \multicolumn{3}{c}{\multirow{1}{*}{$1.71\pm0.40$}} &
			\multicolumn{3}{c}{\multirow{1}{*}{$\textbf{2.19}$~$\pm$~$\textbf{0.23}$}} & 
			\multicolumn{3}{c}{\multirow{1}{*}{$2.53\pm0.62$}}  
        \\ 
			\bottomrule[1pt]
		\end{tabular}
		}  
        \end{minipage}
    \end{minipage}
\subsection{Regression}
\vspace{-1mm}
\textbf{Experimental Setup.} Following~\cite{li2017meta}, we construct the K-shot regression task by sampling from the target sine curve $y(x)\!=\!Asin(\omega x+b)$, where the amplitude $A\!\in\![0.1,5.0]$, the frequency $\omega \!\in\! [0.8,1.2]$, %and
the phase $b\!\in\![0,\pi]$ and $x$ is sampled from $[-5.0, 5.0]$. %During 
In the meta-training phase, each task contains %both K training and testing 
K support and K target (K=10) examples. %hosen from $[-5.0,5.0]$. The mean squared error (MSE) is chosen as the prediction loss.
We adopt mean squared error (MSE) as the loss function.
For the base model $f_\theta$, we adopt a small neural network, which consists of an input layer of size 1, 2 hidden layers of size 40 with ReLU and an output layer of size 1%, as the regressor
. %In optimization part, w
We use one gradient update with a fixed step size $\alpha\!=\!0.01$ in inner loop, and use Adam as the outer-loop optimizer following~\cite{finn2017model,li2017meta}. Moreover, the meta-learner is trained on 240,000 tasks with meta batch-size %of
being 4. %During 
In meta-testing stage %(i.e., performance evaluation)
, we randomly sample 100 sine curves %, each curve contains K samples for training and 100 examples evenly distributed on $[-5.0,5.0]$ for testing. We choose K as 10, and repeat theis testing procedure 100 times to compute the average MSE.
as meta-test tasks, each task containing K support samples and 100 query examples. The data points $x$ in query set are evenly distributed on $[-5.0,5.0]$.
The averaged MSE with 95$\%$ confidence intervals upon these 100 sine curves with K=10, 20, 30 are %listed on
reported in Table~\ref{Regression}. %We summarize the meta-testing performance on various domains of amplitude $A$, frequency $\omega$ and phase $b$ in Table~\ref{Regression-cross}.  
%For the augmentation network, we set the up-sampled ratio as 16.  
We also perform cross-domain experiments by sampling 100 sine curves which have different frequencies, amplitudes or phases from the tasks in meta-training set and report the results in Table~\ref{Regression-cross}. More settings about the up-sampling networks are listed in Appendix C.

\vspace{-3mm}
\begin{figure}[h]
\centering
\subfigure[MLTI]{
\begin{minipage}[t]{0.24\linewidth}
\centering
\includegraphics[width=1.29in, height = 0.91in]{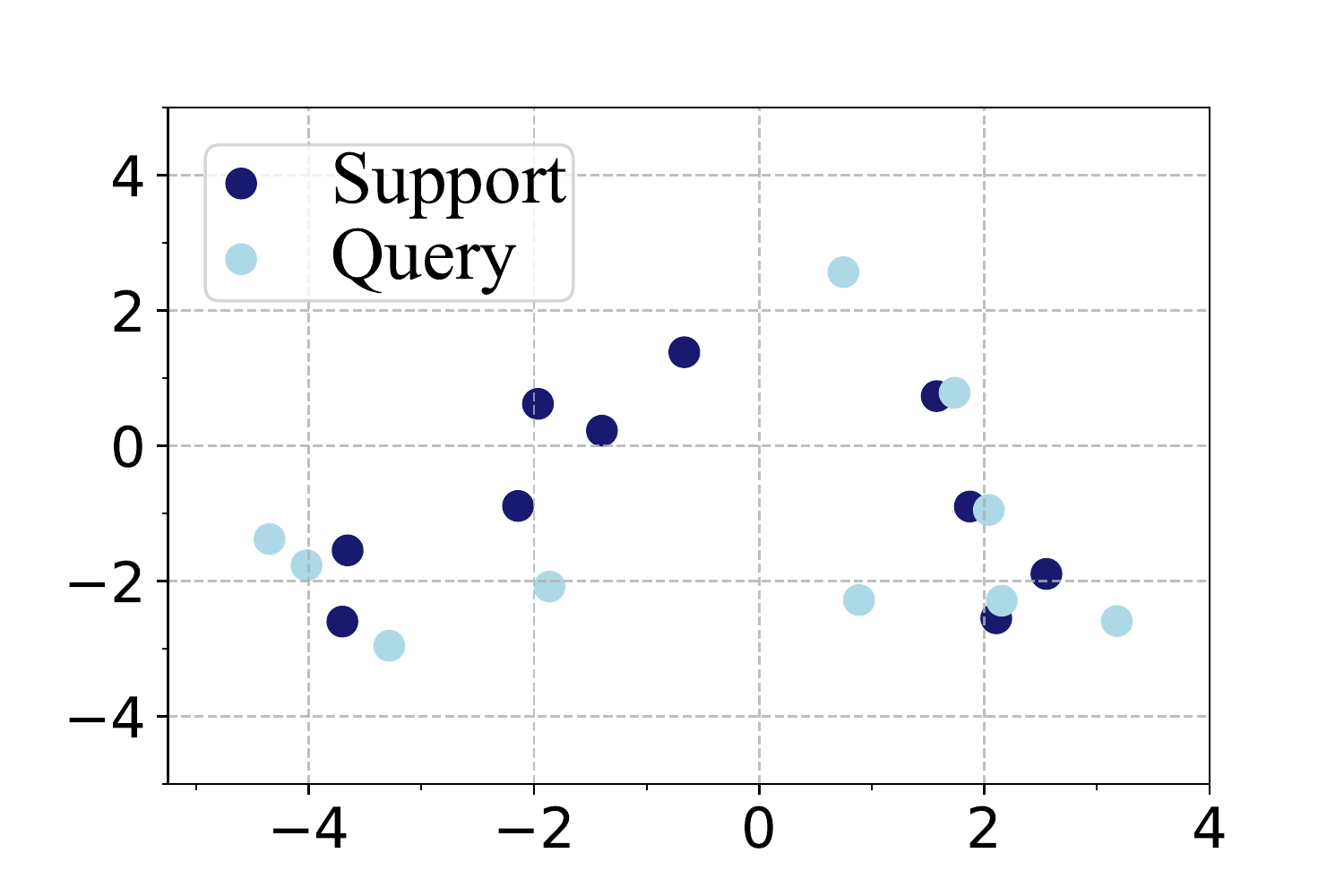}
%\caption{fig1}
\end{minipage}%
}%
\subfigure[MetaMix]{
\begin{minipage}[t]{0.24\linewidth}
\centering
\includegraphics[width=1.29in, height = 0.91in]{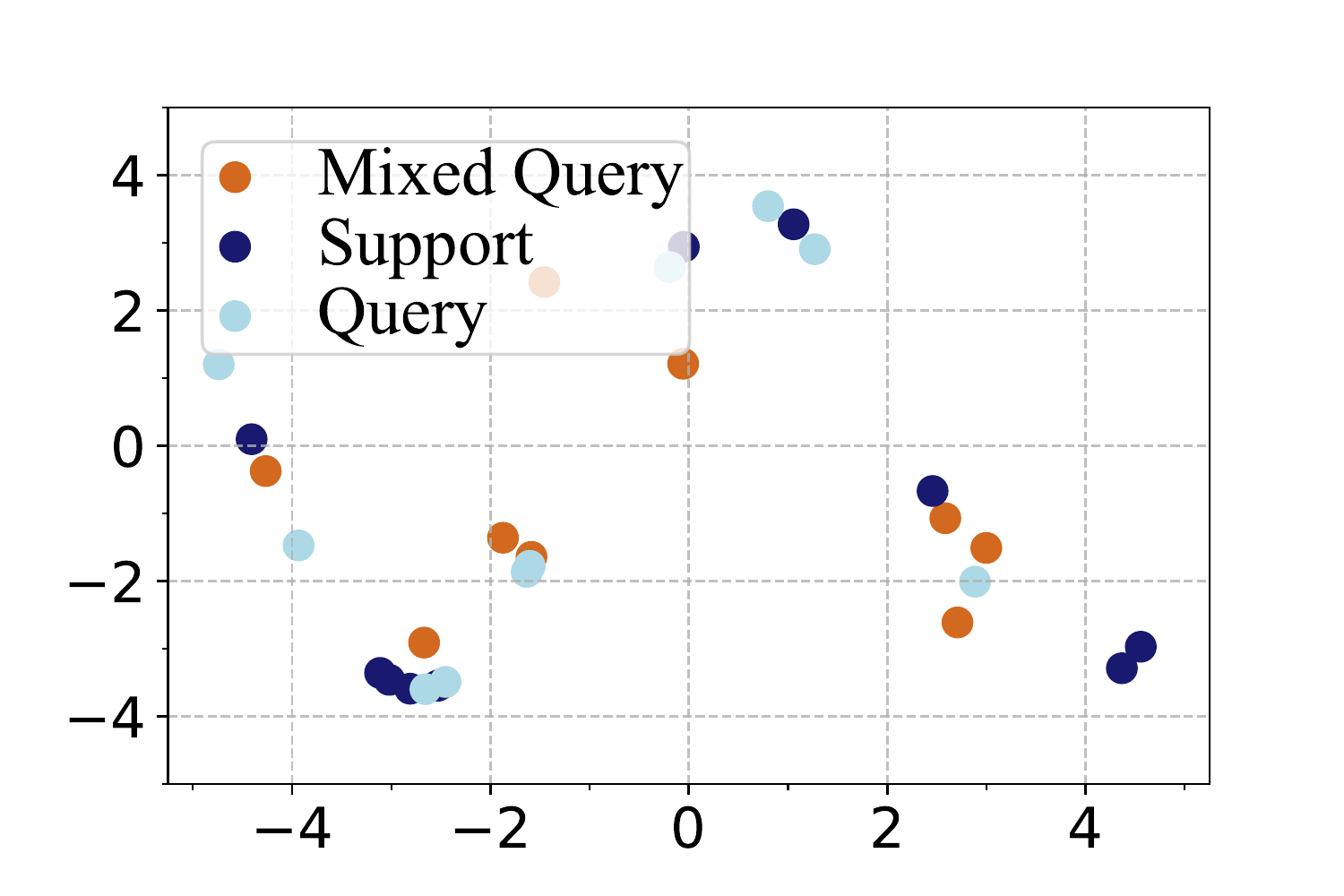}
%\caption{fig1}
\end{minipage}%
}%
\subfigure[%AU
TU]{
\begin{minipage}[t]{0.24\linewidth}
\centering
\includegraphics[width=1.29in, height = 0.91in]{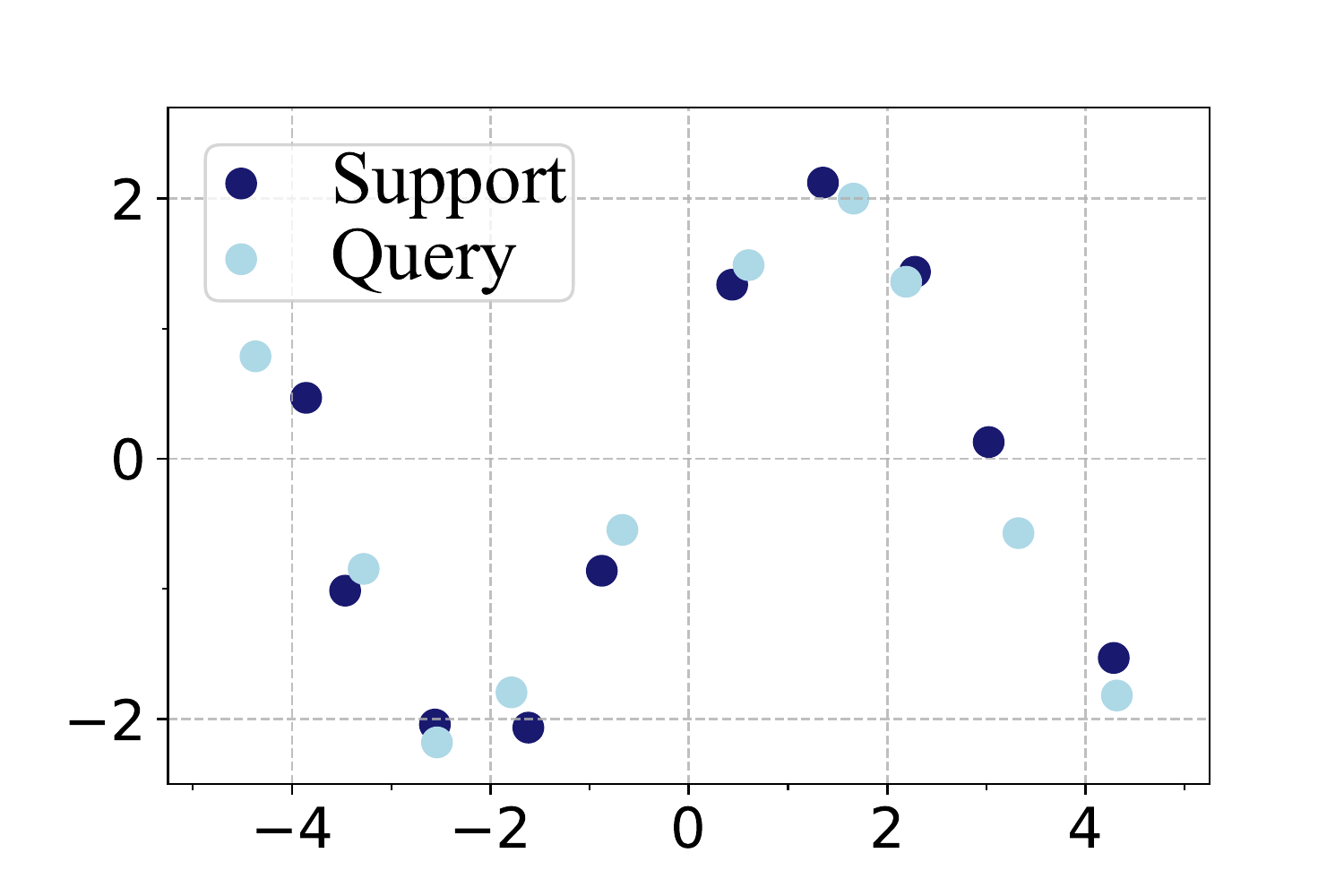}
%\caption{fig2}
\end{minipage}%
}%
\subfigure[%TAU
ATU]{
\begin{minipage}[t]{0.23\linewidth}
\centering
\includegraphics[width=1.29in, height = 0.91in]{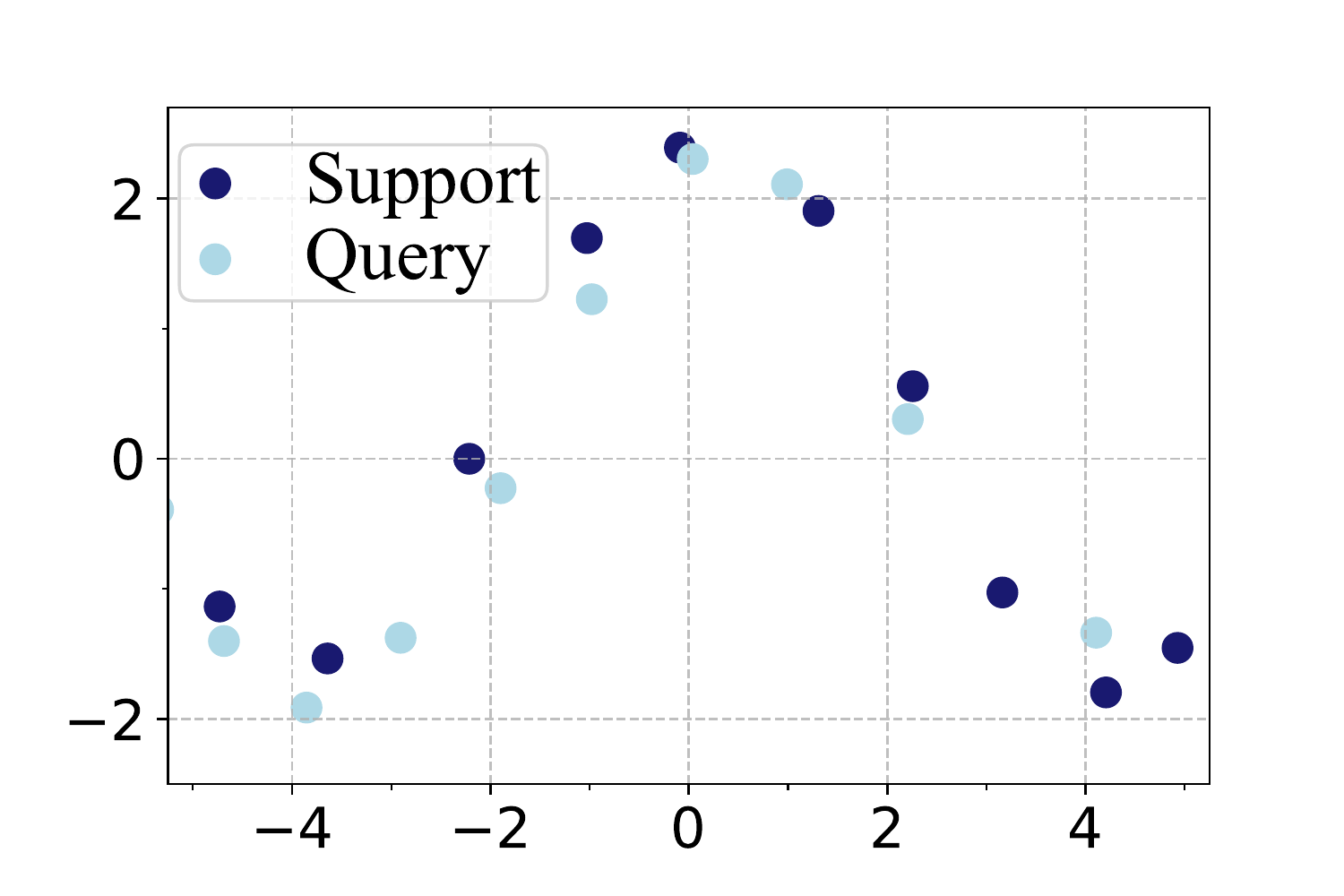}
%\caption{fig2}
\end{minipage}
}%
\centering
\vspace{-2mm}
\caption{The augmented regression tasks generated by different augmentation-based methods.}
\label{generated_regress_task}
\end{figure}
\vspace{-3pt}
\textbf{Performance.}The results in Table~\ref{Regression} and Table~\ref{Regression-cross} show that ATU 
consistently outperforms the baseline methods MAML, MetaMix, and MLTI %under 
in different K-shot (K$\in\{10,20,30\}$) %testing
settings and various domain settings. 
These %outcomes %show 
results validate that the %ATU's generated tasks
tasks generated by ATU can %could 
better %represent
approximate the true task distribution and provide more information to the meta-learner than MetaMix and MLTI, thus enabling better generalization of the model. %and thus empower the model has 
We further verify the superiority of the proposed methods by visualizing the augmented tasks generated by the proposed methods and the baseline methods. %MLTI and MetaMix. 
The visualization results in Fig.~\ref{generated_regress_task} show that the points in the tasks generated by TU and ATU fit the sine curve well, while the points in the tasks 
generated by MLTI and MetaMix deviate from the sine curve. This indicates that augmented tasks generated by TU and ATU match the true task distribution. It is %worth noting 
noteworthy that the support set and query set generated by ATU %are more different than 
differ significantly from those generated by TU, which %means 
indicates that the task generated by ATU is more difficult. This, together with the results that ATU outperforms TU in most experiments, demonstrates the effectiveness of the adversarial losses in generating informative tasks to improve generalization of the meta-learner.

\vspace{-3mm}
\begin{figure}[h]
\centering
\begin{minipage}[b]{0.5\linewidth}

In Fig.~\ref{regression_curves}, we also visualize the adaptation of meta-learner trained by different task augmentation methods for a 10-shot meta-test regression task. Compared to the MAML trained on original meta-training tasks, the MAML trained on tasks generated by TU fits the ground-truth sinusoid after only one update. And ATU performs even better than TU. This again validates that the augmented tasks generated by TU and ATU are more informative for the meta-learner to learn the meta knowledge from the true task distribution.
\end{minipage}
\hfill
\begin{minipage}[b]{0.45\linewidth}

\includegraphics[width=2.15in, height =1.05in]{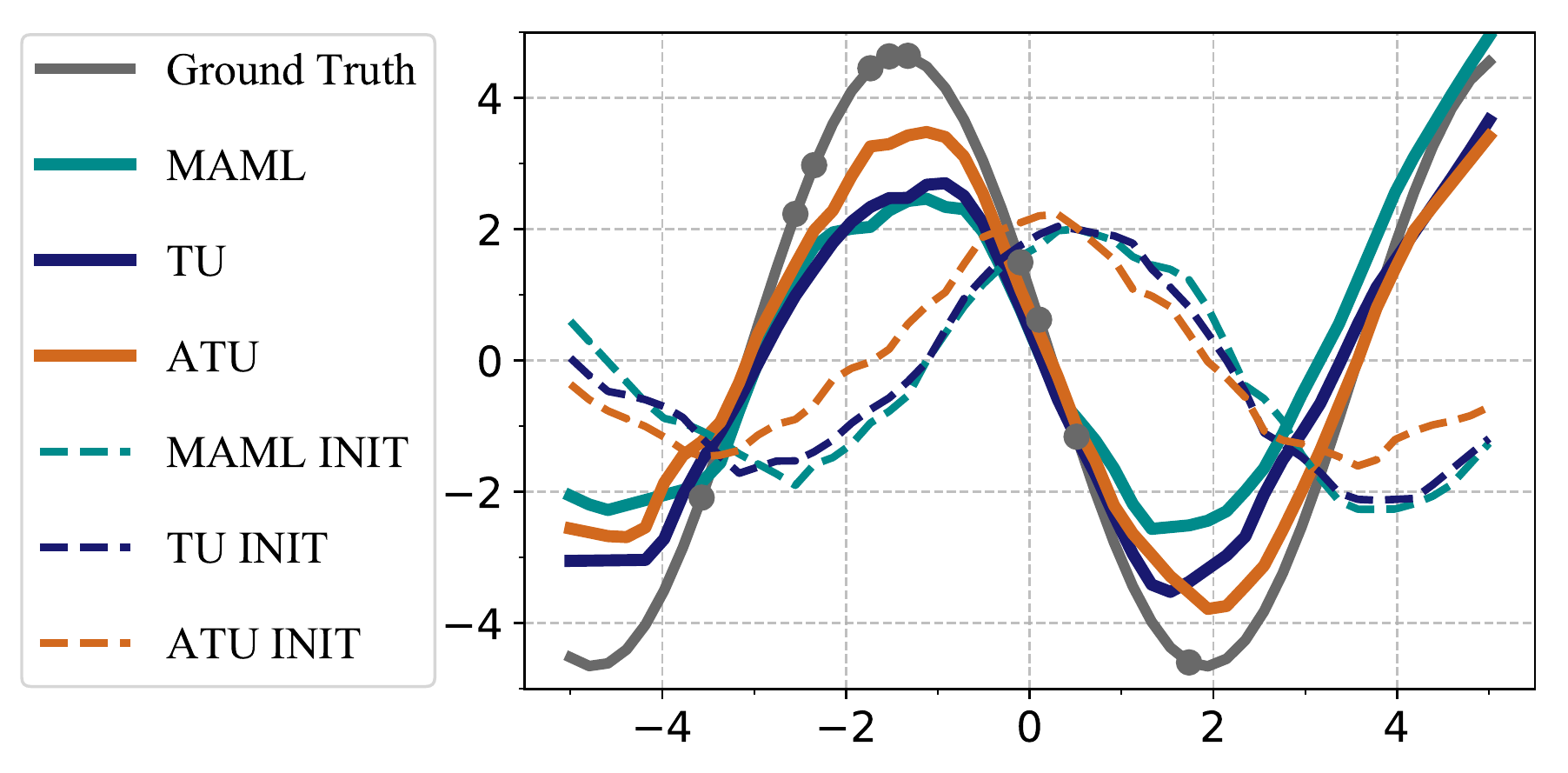}
\caption {%Ten
Initialization (dotted) and one-step adaptation (solid) regression curves of MAML,TU and ATU when K=10.}
\label{regression_curves}
\end{minipage}
\end{figure}

\vspace{-3mm}
\subsection{Classification}
% \vspace{-5mm}
\begin{table*}[h]
    \centering
 	\setlength{\tabcolsep}{1 mm}{
 	\caption{Average accuracy under different settings of few-shot  classification and various datasets.}
 	\label{classification}
 	\vspace{1mm}
 	    \begin{tabular}{ccc|cc|cc|cc|cc|cc|cc|cc|cc|cc} 
			\toprule[1pt]
			\multicolumn{3}{l|}{\multirow{2}{*}{Model}} &
			\multicolumn{2}{c|}{\multirow{1}{*}{miniImagenet-S}} & 
			\multicolumn{2}{c|}{\multirow{1}{*}{ISIC}} &
			\multicolumn{2}{c|}{\multirow{1}{*}{DermNet-S}} &
			\multicolumn{2}{c}{\multirow{1}{*}{Tabular Murris}} \\

			\multicolumn{3}{l|}{\multirow{2}{*}{}} & 
			\multicolumn{1}{c}{\multirow{1}{*}{1-shot}} &
			\multicolumn{1}{c|}{\multirow{1}{*}{5-shot}} &
			\multicolumn{1}{c}{\multirow{1}{*}{1-shot}} &
			\multicolumn{1}{c|}{\multirow{1}{*}{5-shot}} &
			\multicolumn{1}{c}{\multirow{1}{*}{1-shot}} &
			\multicolumn{1}{c|}{\multirow{1}{*}{5-shot}} &
			\multicolumn{1}{c}{\multirow{1}{*}{1-shot}} &
			\multicolumn{1}{c}{\multirow{1}{*}{5-shot}} &\\		
		    \midrule
			\multicolumn{3}{l|}{\multirow{1}{*}{MAML~\cite{finn2017model}}} & 
			\multicolumn{1}{c}{\multirow{1}{*}{$38.27\%$}} &
			\multicolumn{1}{c|}{\multirow{1}{*}{$52.14\%$}} &
			\multicolumn{1}{c}{\multirow{1}{*}{$57.59\%$}} &
			\multicolumn{1}{c|}{\multirow{1}{*}{$65.24\%$}} &
			\multicolumn{1}{c}{\multirow{1}{*}{$43.47\%$}} &
			\multicolumn{1}{c|}{\multirow{1}{*}{$60.56\%$}} &
			\multicolumn{1}{c}{\multirow{1}{*}{$79.08\%$}} &
			\multicolumn{1}{c}{\multirow{1}{*}{$88.55\%$}} \\	 
			
			\multicolumn{3}{l|}{\multirow{1}{*}{Meta-Reg~\cite{yin2020meta}}} & 
			\multicolumn{1}{c}{\multirow{1}{*}{$38.35\%$}} &
			\multicolumn{1}{c|}{\multirow{1}{*}{$51.74\%$}} &
			\multicolumn{1}{c}{\multirow{1}{*}{$58.57\%$}} &
			\multicolumn{1}{c|}{\multirow{1}{*}{$68.45\%$}} &
			\multicolumn{1}{c}{\multirow{1}{*}{$45.01\%$}} &
			\multicolumn{1}{c|}{\multirow{1}{*}{$60.92\%$}} &
			\multicolumn{1}{c}{\multirow{1}{*}{$79.18\%$}} &
			\multicolumn{1}{c}{\multirow{1}{*}{$89.08\%$}} \\	
		
			\multicolumn{3}{l|}{\multirow{1}{*}{TAML~\cite{jamal2019task}}} & 
			\multicolumn{1}{c}{\multirow{1}{*}{$38.70\%$}} &
			\multicolumn{1}{c|}{\multirow{1}{*}{$52.75\%$}} &
			\multicolumn{1}{c}{\multirow{1}{*}{$58.39\%$}} &
			\multicolumn{1}{c|}{\multirow{1}{*}{$66.09\%$}} &
			\multicolumn{1}{c}{\multirow{1}{*}{$45.73\%$}} &
			\multicolumn{1}{c|}{\multirow{1}{*}{$61.14\%$}} &
			\multicolumn{1}{c}{\multirow{1}{*}{$79.82\%$}} &
			\multicolumn{1}{c}{\multirow{1}{*}{$89.11\%$}} \\	

			\multicolumn{3}{l|}{\multirow{1}{*}{Meta-Dropout~\cite{lee2019meta}}} & 
			\multicolumn{1}{c}{\multirow{1}{*}{$38.32\%$}} &
			\multicolumn{1}{c|}{\multirow{1}{*}{$52.53\%$}} &
			\multicolumn{1}{c}{\multirow{1}{*}{$58.40\%$}} &
			\multicolumn{1}{c|}{\multirow{1}{*}{$67.32\%$}} &
			\multicolumn{1}{c}{\multirow{1}{*}{$44.30\%$}} &
			\multicolumn{1}{c|}{\multirow{1}{*}{$60.86\%$}} &
			\multicolumn{1}{c}{\multirow{1}{*}{$78.18\%$}} &
			\multicolumn{1}{c}{\multirow{1}{*}{$89.25\%$}} \\	

			\multicolumn{3}{l|}{\multirow{1}{*}{MetaMix~\cite{yao2021improving}}} & 
			\multicolumn{1}{c}{\multirow{1}{*}{$39.43\%$}} &
			\multicolumn{1}{c|}{\multirow{1}{*}{$54.14\%$}} &
			\multicolumn{1}{c}{\multirow{1}{*}{$60.34\%$}} &
			\multicolumn{1}{c|}{\multirow{1}{*}{$69.47\%$}} &
			\multicolumn{1}{c}{\multirow{1}{*}{$46.81\%$}} &
			\multicolumn{1}{c|}{\multirow{1}{*}{$63.52\%$}} &
			\multicolumn{1}{c}{\multirow{1}{*}{$81.06\%$}} &
			\multicolumn{1}{c}{\multirow{1}{*}{$89.75\%$}} \\
			
			\multicolumn{3}{l|}{\multirow{1}{*}{Meta-Maxup~\cite{ni2021data}}} & 
			\multicolumn{1}{c}{\multirow{1}{*}{$39.28\%$}} &
			\multicolumn{1}{c|}{\multirow{1}{*}{$53.02\%$}} &
			\multicolumn{1}{c}{\multirow{1}{*}{$58.68\%$}} &
			\multicolumn{1}{c|}{\multirow{1}{*}{$69.16\%$}} &
			\multicolumn{1}{c}{\multirow{1}{*}{$46.10\%$}} &
			\multicolumn{1}{c|}{\multirow{1}{*}{$62.64\%$}} &
			\multicolumn{1}{c}{\multirow{1}{*}{$79.56\%$}} &
			\multicolumn{1}{c}{\multirow{1}{*}{$88.88\%$}} \\

			\multicolumn{3}{l|}{\multirow{1}{*}{MLTI~\cite{yao2021meta}}} & 
			\multicolumn{1}{c}{\multirow{1}{*}{$41.58\%$}} &
			\multicolumn{1}{c|}{\multirow{1}{*}{$55.22\%$}} &
			\multicolumn{1}{c}{\multirow{1}{*}{$61.79\%$}} &
			\multicolumn{1}{c|}{\multirow{1}{*}{$70.69\%$}} &
			\multicolumn{1}{c}{\multirow{1}{*}{$48.03\%$}} &
			\multicolumn{1}{c|}{\multirow{1}{*}{$64.55\%$}} &
			\multicolumn{1}{c}{\multirow{1}{*}{$81.73\%$}} &
			\multicolumn{1}{c}{\multirow{1}{*}{$91.08\%$}} \\
            
            \midrule
			\multicolumn{3}{l|}{\multirow{1}{*}{TU}} & 
			\multicolumn{1}{c}{\multirow{1}{*}{$42.16\%$}} &
			\multicolumn{1}{c|}{\multirow{1}{*}{$56.33\%$}} &
			\multicolumn{1}{c}{\multirow{1}{*}{$62.03\%$}} &
			\multicolumn{1}{c|}{\multirow{1}{*}{$73.97\%$}} &
			\multicolumn{1}{c}{\multirow{1}{*}{$48.07\%$}} &
			\multicolumn{1}{c|}{\multirow{1}{*}{$64.81\%$}} &
			\multicolumn{1}{c}{\multirow{1}{*}{$81.88\%$}} &
			\multicolumn{1}{c}{\multirow{1}{*}{$91.15\%$}} \\	

			\multicolumn{3}{l|}{\multirow{1}{*}{ATU}} & 
			\multicolumn{1}{c}{\multirow{1}{*}{$\textbf{42.60}\%$}} &
			\multicolumn{1}{c|}{\multirow{1}{*}{$\textbf{56.78}\%$}} &
			\multicolumn{1}{c}{\multirow{1}{*}{$\textbf{62.84}\%$}} &
			\multicolumn{1}{c|}{\multirow{1}{*}{$\textbf{74.50}\%$}} &
			\multicolumn{1}{c}{\multirow{1}{*}{$\textbf{48.33}\%$}} &
			\multicolumn{1}{c|}{\multirow{1}{*}{$\textbf{65.16}\%$}} &
			\multicolumn{1}{c}{\multirow{1}{*}{$\textbf{82.04}\%$}} &
			\multicolumn{1}{c}{\multirow{1}{*}{$\textbf{91.42}\%$}} \\
			\bottomrule[1pt]
		\end{tabular}
		}   
\end{table*}
% \vspace{-2mm}
\textbf{Experimental setup.} 
We follow MLTI to evaluate the performance of task augmentation algorithms for few-shot classification problem with limited number of base classes in meta-training set under non-label-sharing settings. We consider
four datasets (base classes number): miniImagenet-S (12), ISIC~\cite{milton2019automated} (4), Dermnet-S (30), and Tabular Murris~\cite{cao2020concept} (57) %, which include 
covering classification tasks on general natural images, medical images, and gene data. 
Note that the miniImagenet-S and Dermnet-S are constructed by limiting the base classes of miniImangenet~\cite{vinyals2016matching} and Dermnet~\cite{Dermnet}, respectively. %For the 
We construct $N$-way $K$-shot %setting,
tasks, setting %we choose
$N=5$ for miniImagenet-S, Derment-S, Tabular Murris and %let 
$N=2$ for ISIC dataset due to its limited number of base classes and setting $K=1$ or $K=5$. Recall that TAU relies on extra $K_M$ image to generate augmented tasks in image classification problem. To make the training process more efficient, we set $K_{M}$ to 3 and the up-sampling ratio $r$ be 2. More details of these datasets and the settings of the task augmentation networks are listed in Appendix D.

\textbf{Performance.} We show the performance on the four datasets in Table~\ref{classification}. On all four datasets, the proposed ATU consistently outperforms the baseline methods, including the augmentation-based methods (i.e. MetaMix, Meta-Maxup and MLTI) and regularization-based methods (Meta-Reg, TAML and Meta-Dropout). And TU achieves the second best performance on all experiments.  We also observe that our method achieves a large improvement on the ISIC dataset which consists of only 4 base classes, indicating the effectiveness of our method in limited tasks scenarios. 
We further evaluate the effectiveness of the proposed ATU on improving the generalization for different backbone meta-learner by conducting experiments under 1-shot setting to compare the performance of MLTI and ATU in improving the performance of the backbone meta-learner MetaSGD and ANIL. The results are presented in Table~\ref{Classification_backbones}. ATU again consistently outperforms MLTI.
All these results validate the superiority of the proposed ATU and TU over the existing baselines in generating informative tasks to improve the performance of different backbone meta-learners. 
We also evaluate the performance of ATU in cross-domain adaptaion settings. In Table~\ref{classification_cross}, we present the results of the experiment that apply the meta-model trained on miniImageNet-S to Dermnet-S, and vice versa. ATU improves the generalization performance of MAML (the backbone meta-learner in this experiment) by a large margin. This indicates ATU can consistently improve the backbone meta-model's generalization ability under the challenging cross-domain settings.
% \vspace{-6pt}
\begin{minipage}[t!]{\textwidth}
        \begin{minipage}[b]{0.4\textwidth}
        \centering
    \makeatletter\def\@captype{table}
    \caption{%Compatibility analysis on different meta-learning algorithms
    Comparison of compatibility with different backbone meta-learning algorithms on 1-shot classification.}
    \label{Classification_backbones}
        \resizebox{66.5mm}{13mm}{
 	    \begin{tabular}{ccc|cc|cc|cc|cc} 
		%\begin{tabular}{ccc|ccc|ccc} 
			\toprule
			\multicolumn{3}{l}{\multirow{1}{*}{Method}} &
% 			\multicolumn{2}{c|}{\multirow{1}{*}{Strategies}} & 
			\multicolumn{2}{c|}{\multirow{1}{*}{mini-S}} & 
			\multicolumn{2}{c|}{\multirow{1}{*}{ISIC}} & 
			\multicolumn{2}{c|}{\multirow{1}{*}{Derm-S}} & 
			\multicolumn{2}{c}{\multirow{1}{*}{TM}} \\
			
			\midrule
		    %\multicolumn{1}{c}{\multirow{1}{*}{\qquad\textit{MetaSGD as backbone}}}   \\ 
			\multicolumn{3}{l}{\multirow{1}{*}{MetaSGD~\cite{li2017meta}}} &
% 			\multicolumn{2}{c|}{\multirow{1}{*}{}} & 
			\multicolumn{2}{c|}{\multirow{1}{*}{$37.88\%$}} & 
			\multicolumn{2}{c|}{\multirow{1}{*}{$58.79\%$}} &
			\multicolumn{2}{c|}{\multirow{1}{*}{$42.07\%$}} & 
			\multicolumn{2}{c}{\multirow{1}{*}{$81.55\%$}}  \\

			\multicolumn{3}{l}{\multirow{1}{*}{MetaSGD+MLTI}} &
% 			\multicolumn{2}{c|}{\multirow{1}{*}{+MLTI}} & 
			\multicolumn{2}{c|}{\multirow{1}{*}{$39.58\%$}} & 
			\multicolumn{2}{c|}{\multirow{1}{*}{$61.57\%$}} & 
			\multicolumn{2}{c|}{\multirow{1}{*}{$45.49\%$}} & 
			\multicolumn{2}{c}{\multirow{1}{*}{$83.31\%$}}  \\

			\multicolumn{3}{l}{\multirow{1}{*}{MetaSGD+ATU}} &
% 			\multicolumn{2}{c|}{\multirow{1}{*}{+ATU}} & 
			\multicolumn{2}{c|}{\multirow{1}{*}{$\textbf{40.52\%}$}} & 
			\multicolumn{2}{c|}{\multirow{1}{*}{$\textbf{62.84\%}$}} & 
			\multicolumn{2}{c|}{\multirow{1}{*}{$\textbf{46.78\%}$}} & 
			\multicolumn{2}{c}{\multirow{1}{*}{$\textbf{83.84\%}$}}  \\
            
			\midrule
		    %\multicolumn{1}{c}{\multirow{1}{*}{\qquad\textit{ANIL as backbone}}}   \\ 			
			
			\multicolumn{3}{l}{\multirow{1}{*}{ANIL~\cite{raghu2019rapid}}} &
% 			\multicolumn{2}{c|}{\multirow{1}{*}{}} & 
			\multicolumn{2}{c|}{\multirow{1}{*}{$38.02\%$}} & 
			\multicolumn{2}{c|}{\multirow{1}{*}{$59.48\%$}} &
			\multicolumn{2}{c|}{\multirow{1}{*}{$44.58\%$}} & 
			\multicolumn{2}{c}{\multirow{1}{*}{$75.67\%$}}  \\

			\multicolumn{3}{l}{\multirow{1}{*}{ANIL+MLTI}} &
% 			\multicolumn{2}{c|}{\multirow{1}{*}{+MLTI}} & 
			\multicolumn{2}{c|}{\multirow{1}{*}{$39.15\%$}} & 
			\multicolumn{2}{c|}{\multirow{1}{*}{$61.78\%$}} & 
			\multicolumn{2}{c|}{\multirow{1}{*}{$46.79\%$}} & 
			\multicolumn{2}{c}{\multirow{1}{*}{$77.11\%$}}  \\

			\multicolumn{3}{l}{\multirow{1}{*}{ANIL+ATU}} &
% 			\multicolumn{2}{c|}{\multirow{1}{*}{+ATU}} & 
			\multicolumn{2}{c|}{\multirow{1}{*}{$\textbf{39.27\%}$}} & 
			\multicolumn{2}{c|}{\multirow{1}{*}{$\textbf{62.12\%}$}} & 
			\multicolumn{2}{c|}{\multirow{1}{*}{$\textbf{47.03\%}$}} & 
			\multicolumn{2}{c}{\multirow{1}{*}{$\textbf{77.23\%}$}}  \\	
			\bottomrule
		\end{tabular}
		}   
        \end{minipage}
        \hspace{0.4in}
    \begin{minipage}[b]{0.5\textwidth}

        \centering
    \makeatletter\def\@captype{table}
	\caption{Cross-domain adaptation experiments between mini-S and Dermnet-S. A$\!\rightarrow$B denotes that the %base model
	backbone meta-model is meta-trained on A and meta-tested on B.}
        \label{classification_cross}
        \resizebox{72mm}{13 mm}{ %17.4mm as backbone
 	    \begin{tabular}{ccc|cccc|cccc} 
			\toprule
			\multicolumn{3}{l}{\multirow{2}{*}{Model}} &
			\multicolumn{4}{c|}{\multirow{1}{*}{mini-S$\rightarrow$ Derm-S}} &
			\multicolumn{4}{c}{\multirow{1}{*}{Derm-S$\rightarrow$ mini-S}} \\
			\multicolumn{3}{l}{\multirow{1}{*}{}} & 
% 			\multicolumn{2}{c|}{\multirow{1}{*}{}} &
			\multicolumn{2}{c}{\multirow{1}{*}{1-shot}} &
			\multicolumn{2}{c|}{\multirow{1}{*}{5-shot}} &
			\multicolumn{2}{c}{\multirow{1}{*}{1-shot}} &
			\multicolumn{2}{c}{\multirow{1}{*}{5-shot}}\\	
            
            \midrule
		    %\multicolumn{1}{c}{\multirow{1}{*}{\qquad %\textit{MAML as backbone}}}   \\ 
			\multicolumn{3}{l}{\multirow{1}{*}{MAML~\cite{finn2017model}}} & 
% 			\multicolumn{2}{l|}{\multirow{1}{*}{}} &
			\multicolumn{2}{c}{\multirow{1}{*}{$34.46\%$}} &
			\multicolumn{2}{c|}{\multirow{1}{*}{$50.36\%$}} &
			\multicolumn{2}{c}{\multirow{1}{*}{$28.78\%$}} &
			\multicolumn{2}{c}{\multirow{1}{*}{$41.29\%$}}  \\

			\multicolumn{3}{l}{\multirow{1}{*}{MAML+ATU}} & 
% 			\multicolumn{2}{l|}{\multirow{1}{*}{+ATU}} &
			\multicolumn{2}{c}{\multirow{1}{*}{$\textbf{36.86}\%$}} &
			\multicolumn{2}{c|}{\multirow{1}{*}{$\textbf{51.98}\%$}} &
			\multicolumn{2}{c}{\multirow{1}{*}{$\textbf{30.68}\%$}} &
			\multicolumn{2}{c}{\multirow{1}{*}{$\textbf{46.72}\%$}}  \\		    

            \midrule
		    %\multicolumn{1}{c}{\multirow{1}{*}{\qquad \textit{MetaSGD as backbone}}}   \\ 
			\multicolumn{3}{l}{\multirow{1}{*}{MetaSGD~\cite{li2017meta}}} & 
% 			\multicolumn{2}{l|}{\multirow{1}{*}{}} &
			\multicolumn{2}{c}{\multirow{1}{*}{$31.07\%$}} &
			\multicolumn{2}{c|}{\multirow{1}{*}{$49.07\%$}} &
			\multicolumn{2}{c}{\multirow{1}{*}{$28.17\%$}} &
			\multicolumn{2}{c}{\multirow{1}{*}{$41.83\%$}}  \\		
			
			\multicolumn{3}{l}{\multirow{1}{*}{MetaSGD+ATU}} & 
% 			\multicolumn{2}{l|}{\multirow{1}{*}{ATU}} &
			\multicolumn{2}{c}{\multirow{1}{*}{$\textbf{37.75\%}$}} &
			\multicolumn{2}{c|}{\multirow{1}{*}{$\textbf{54.60\%}$}} &
			\multicolumn{2}{c}{\multirow{1}{*}{$\textbf{30.78\%}$}} &
			\multicolumn{2}{c}{\multirow{1}{*}{$\textbf{44.01\%}$}}  \\	
			\bottomrule
		\end{tabular}
		}   
% 	\caption{Cross-domain adaptation experiments between mini-S and Dermnet-S. A$\!\rightarrow$B denotes that the %base model
% 	\kai{backbone meta-model} is meta-trained on A and meta-tested on B.}
%         \label{classification_cross}
        \end{minipage}
    \end{minipage}

\begin{figure}[h]
\begin{minipage}[t]{.5\linewidth}
    \vspace{0pt}
    \centering
    \includegraphics[width=2.15in, height = 1.35in]{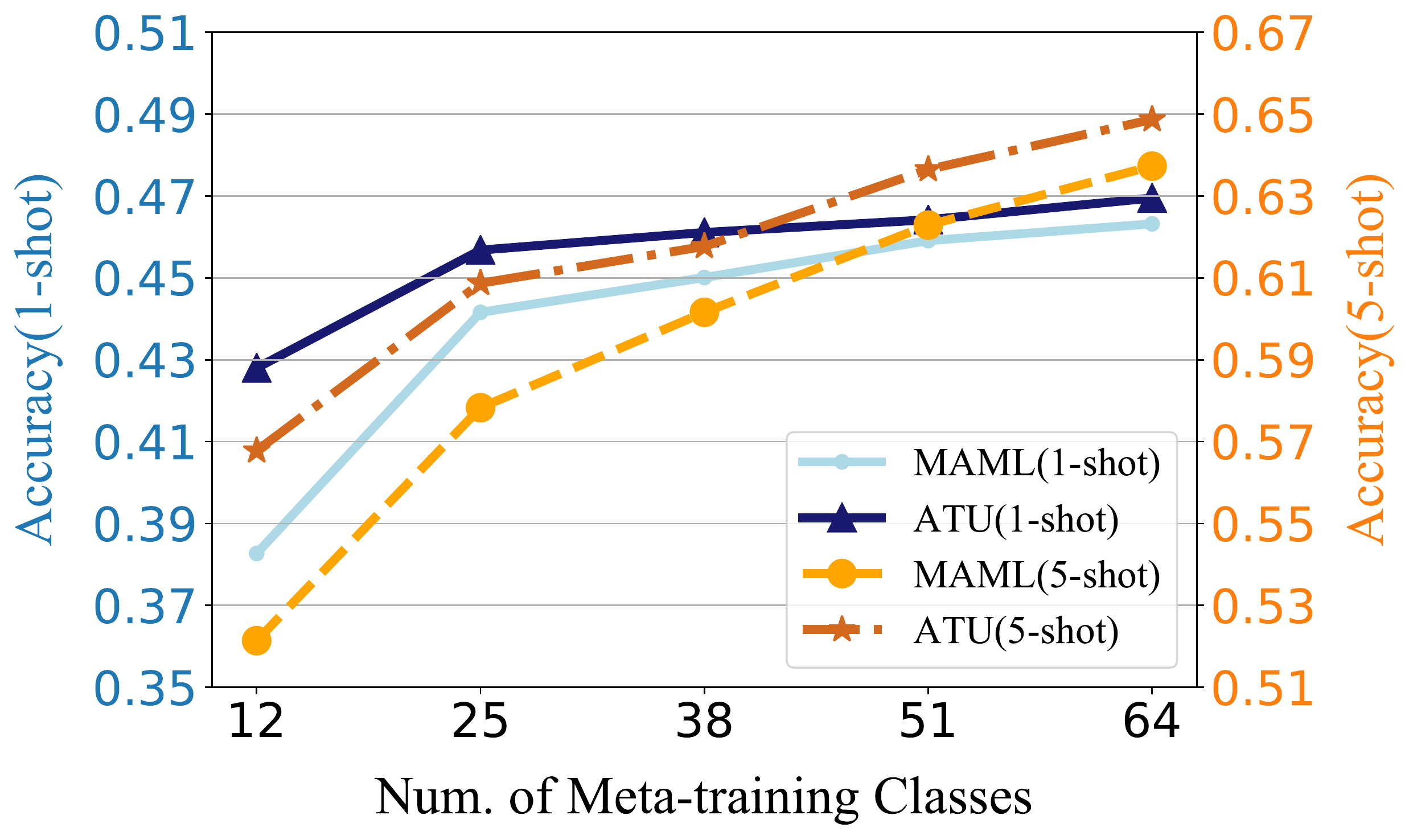}
    \caption{The averaged accuracy on the miniImagenet dataset with different number of tasks.}
    \label{task_num}
\end{minipage}
\hspace{0.05in}
\begin{minipage}[t]{.5\linewidth}
    \vspace{0pt}
    \centering
    \includegraphics[width=1.85 in, height = 1.35in]{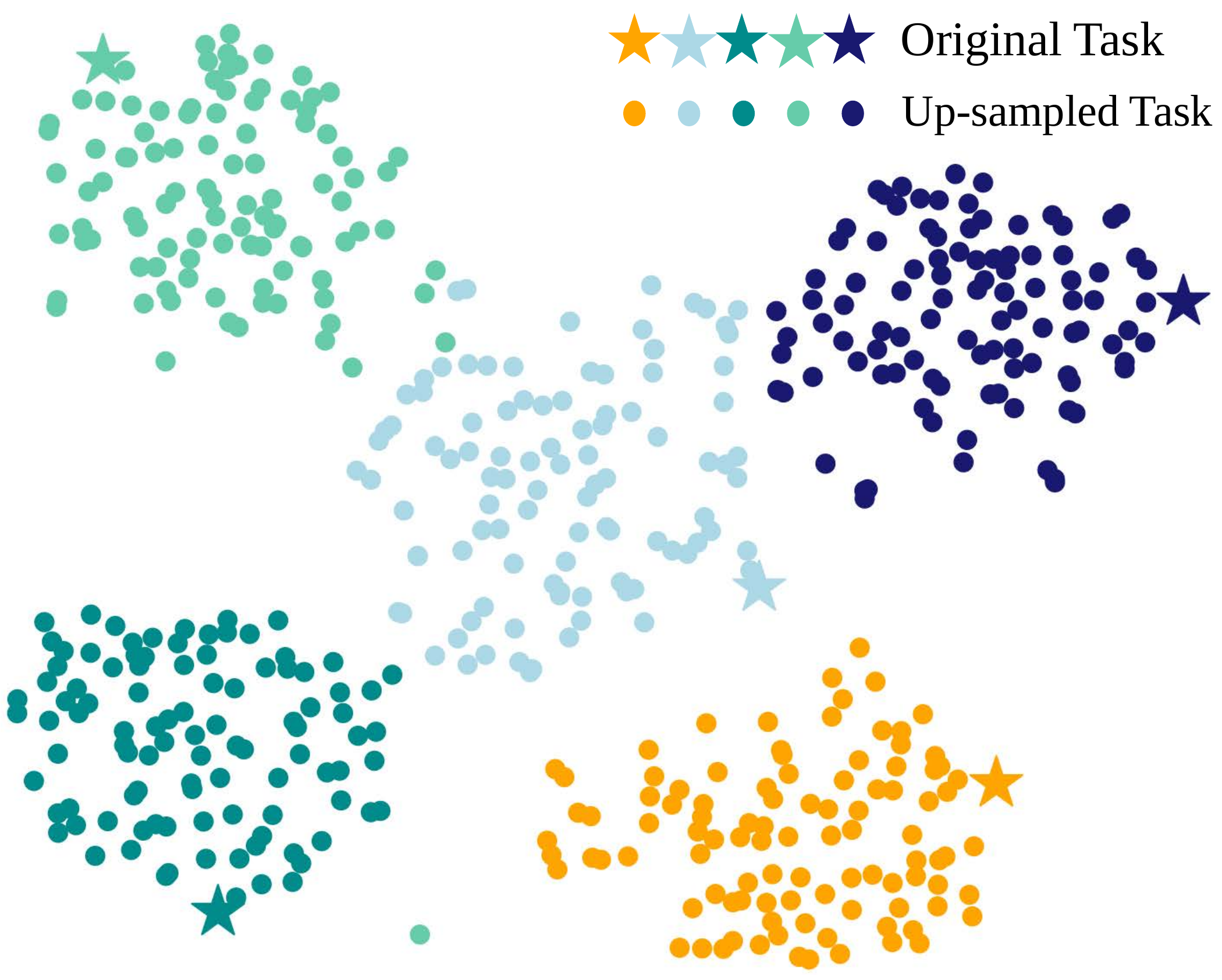}
    \caption{T-SNE visualization of original and up-sampled tasks on 1-shot miniImagenet-S setting.}
    \label{tsne}
\end{minipage}
\end{figure}

\vspace{10pt}
\textbf{Effect of the number of meta-training tasks.} We analyze the change of the performance improvement of ATU's over MAML with the number of meta-training tasks in 1-shot and 5-shot settings. The results presented in  Fig.~\ref{task_num} shows that ATU significantly improve the performance of MAML by about 4.5\%-5\% %in 1-shot and 5-shot settings 
when the number of base classes is 12, while the improvement %Then the performance improvement 
decreases with the number of base classes increasing. When the number of base classes increases, the number of training tasks increases rapidly and the empirical task distribution constructed from the meta-training tasks becomes closer and closer to the true task distribution. Therefore, the extra information provided by tasks generated by ATU becomes less. However, even if all available base classes are used in the meta-training (i.e., 64 meta-training classes), ATU still helps to improve the performance of MAML.

\textbf{Visualization of the generated tasks.} We visualize the up-sampled tasks by t-SNE to evaluate their quality for MAML on miniImagenet-S. We up-sample 100 tasks for 5 original tasks via ATU by using different perturbations for each task. Each task is represented by the concatenated vector of the support and query sets. The results presented in Fig.~\ref{tsne} shows that the up-sampled tasks stays near the original tasks, matching the true task distribution. This indicates the generated tasks are task-aware. Besides, we can observe that the augmented tasks are diverse and cover a substantial portion around the original tasks. This demonstrates the task imaginary property of the augmented tasks. These two observations suggest that the proposed ATU is a qualified task augmentation algorithm.
 
\label{sec:experiments}

\vspace{-2mm}
\section{Conclusion and Limitation}
\vspace{-3mm}

In this paper, we propose the first task-level up-sampling network that learns to generate tasks that simultaneously meet the qualifications of being task-aware, task-imaginary, and model-adaptive. The proposed Adversarial Task Up-sampling (ATU) takes a set of tasks as input and learn to up-sample tasks complying with the true task distribution while being informative to improve the generalization of the meta-learner. We theoretically justify that ATU promotes task-awareness and empirically verify that ATU improves the generalization of various backbone meta-learner for both regression and classification tasks on five datasets. \textbf{Limitations.} Our theoretical results are obtained under some strong assumptions, but all the experiments and visualization outcomes validate our method is effective and indeed could improve the meta learner's generalization in real settings.

\label{sec:conclusion}

% \clearpage
% \bibliographystyle{plain}
\bibliographystyle{plainnat}
\bibliography{main.bib}
\appendix

\clearpage
%\tableofcontents
\setcounter{page}{1}
\setcounter{table}{6}
\setcounter{figure}{6}
\clearpage
\section{Pseudo-codes}

We present the pseudo-codes for the task upsampling network $g(\cdot)$ (Algorithm~\ref{alg:up_sample}), and the meta-training algorithm for the regression task (Algorithm~\ref{alg:regress_meta_train}) and the classification task (Algorithm~\ref{alg:classification_meta_train}) in 1-step MAML with ATU.
For regression task, we randomly sample a batch of tasks as the ground-truth task set $\mathcal{T}_g$ and construct the task patch by down-sampling (FPS sampling). 
For classification tasks, we construct ($K^s\! +\! K^q$) tasks in one shot to obtain %the 
a task patch from a $K^s$-shot classification task with $K^q$ query samples. 
We assume the local task distribution %are
to be smooth, and construct the ground-truth tasks by perform mixup for each image in each task with a nearest image in $K_M$ images. We name the set of $K_M$ images by \emph{memory bank} and denote it by $\mathcal{I}_M$. The $K_M$ images are randomly sampled from classes different from those in the input tasks.
It worth note that ATU is only applied in the meta-training phase and, therefore, the meta-testing phase remains the same as the original 1-step MAML. 
It is direct to extend 1-step MAML to multi-step MAML and extend MAML to ANIL, Meta-SGD  without modifying ATU.

\vspace{2mm}
\begin{algorithm}[h]
    \caption{Task Up-sampling Network $g(\cdot)$}
    \label{alg:up_sample}
    \begin{algorithmic}[1]

        \REQUIRE up-sampling ratio $r=r_c\times r_d$ for the %of 
        coarse generator and decoder, respectively
        % \REQUIRE up-sampling ratio $r_c$ and $r_d$ \ying{for the} %of 
        % coarse generator and decoder\ying{, respectively}
        \STATE \textbf{Input:} a task patch $\mathcal{T}_p = \{T_i\}_{i=1}^{N_p}$
        \STATE Extract the set feature for the input
        task patch $h_s=g_s(\mathcal{T}_p)$  
        \STATE Generate a set of coarse tasks %set 
        $\mathcal{T}_c = g_c(\mathcal{T}_p)$ with set size $r_c N_p$
        \STATE Sample a set of perturbations $\mathcal{Z}$ %with
        in size $r_d$ 
        \STATE Generate the up-sampled task set $\mathcal{T}_{up} = g_d(\mathcal{T}_c, \mathcal{Z}, h_s)$
        \STATE \textbf{Output:} a set of up-sampled tasks $\mathcal{T}_{up}$
    \end{algorithmic}
\end{algorithm}
\vspace{2mm}
\begin{algorithm}[h]

    \caption{Meta-training of 1-step MAML with ATU for regression tasks}
    \label{alg:regress_meta_train}
    \begin{algorithmic}[1]
      % randomly initialized the model's parameters $\theta$;
        \REQUIRE %task
        distribution over meta-training tasks $p(\mathcal{T})$; inner-loop and outer-loop %and outer-loop 
        learning rates %$\alpha,\beta$
        $\alpha$,$\beta$; hyperparameters $\eta_1,\eta_2,\eta_3$ in Eq.~(4) and Eq.~(6); %\kai{meta} 
        batch size of tasks $B$; %B;  %pre-trained up-sampling network $g(\cdot)$.
        task patch size $N_p$; up-sampling ratio $r=r_c\times r_d$ for the task up-sampling network
        % \FOR{epochs = $1$ to $N$}
            \STATE Randomly initialize the parameters $\theta_0$ of the meta-model.
            \WHILE{not converge}
                \STATE Randomly sample a batch of tasks as $\mathcal{T}_g$ with batch size $rN_p$
                %\STATE Down-sampling the batch of tasks to get the local task patch $\mathcal{T}_p$ \kai{with set size $N_p$}.
                \STATE Perform down-sampling (FPS sampling) on $\mathcal{T}_g$ to construct the local task patch $\mathcal{T}_p$
                \STATE %Up-sampled tasks from the task patch and get the augmentation tasks $\mathcal{T}_{up}=g(\mathcal{T}_{p})$.
                Generate the %up-sampling
                augmented task set through our Task Up-sampling Network as $\mathcal{T}_{up} = g(\mathcal{T}_{p})$
                \STATE Randomly split the up-sampled task set $\mathcal{T}_{up}$ into $n$ batches $\{\mathcal{T}_{batch}\}$, each with $B$ tasks (i.e., $n=|\mathcal{T}_{up}|/B$)
                \FOR{each task  %$\hat{T}_i \in \{\mathcal{T}_{up}\}_{j=1}^{n}$ 
                batch $\mathcal{T}_{batch}$ in $\mathcal{T}_{up}$}
                    \FOR{each task $T_i \in \mathcal{T}_{batch}$}
                        \STATE %Calculate the task-specific parameters 
                        Perform inner-loop update of MAML as $\phi_i=\theta_0-\alpha\nabla_{\theta_0}\mathcal{L}(f_{\theta_0},{D}^{s}_{i})$
                    \ENDFOR
                    \STATE Calculate $\mathcal{L}(f_{\phi_{i}},{D}_{i}^{q})$ and $\mathcal{L}_{adv}(\theta_0, D_i^s, D_i^q)$
                    \STATE Update the %initial parameters $\theta_0$ of the model
                    meta-model parameter $\theta_0$ as $\theta_0 \leftarrow \theta_0 - \beta\frac{1}{B}\sum_{i=1}^{n}\nabla_{\theta_0} \mathcal{L}(f_{\phi_{i}},{D}_{i}^{q})$
                \ENDFOR
                \STATE Calculate the objective function in Eq.~(6) and perform backpropagation to update the Task Up-sampling Network
            \ENDWHILE
        % \ENDFOR
    \end{algorithmic}
\end{algorithm}
\vspace{-5mm}

\begin{algorithm}[h]
    \caption{Meta-training of 1-step MAML with ATU for classification tasks (N-way $K^s$-shot)}
    \label{alg:classification_meta_train}
    \begin{algorithmic}[1]
        % \REQUIRE distribution over meta-training tasks $p(\mathcal{T})$; $inner-loop$ and $outer-loop$ learning rates $\alpha$,$\beta$; hyperparameters $\eta_{1}$,$\eta_{2}$ in Eq.~(4); batch size of tasks $B$; task patch size $N_p$;  Beta distribution $Beta(\delta_1,\delta_2)$; up-sampling ratio r ($r=r_c$\times$r_d, r_c=1$)
        \REQUIRE distribution over meta-training tasks $p(\mathcal{T})$; inner-loop and outer-loop learning rates $\alpha$,$\beta$; hyperparameters $\eta_{1}$,$\eta_{2}$ in Eq.~(4); batch size of tasks $B$; task patch size $N_p$;  Beta distribution $Beta(\delta_1,\delta_2)$; up-sampling ratio r ($r=r_c \times r_d, r_c=1$)
        \STATE Randomly initialize the parameters $\theta_0$ of the meta model
        \WHILE{not converge}
        \STATE Randomly sample a batch of tasks $\mathcal{T}_{batch}$ with $B$ tasks.%$\{T_{i}\}_{i=1}^{B}$.\\
        \FOR{each task $T_i \in \mathcal{T}_{batch}$}
            \STATE Reshape $T_i$ as the task patch $\mathcal{T}_p$
            \STATE Randomly sample extra $K_{M}$ images which consists of images not belong to any class in $T_i$
            \STATE Construct the $\mathcal{T}_g = (\hat{C}_{0},...,\hat{C}_{N}):$ Sample $\lambda \sim Beta(\delta_1,\delta_2)$. For the image in each class $C_j$ %of 
            in $T_i$, generate a new image as $\hat{C}_{j}=\lambda\times C_{j} +(1-\lambda)\times X_j$, where $X_j$ %$X_j \in \{K_{M}\}$ 
            is the nearest image (%i.e., 
            measured by euclidean distance) to the image in the class $C_j$. 
            \STATE Generate up-sampling task set through Task Up-sampling Network as $\mathcal{T}_{up} = g(\mathcal{T}_{p})$
            \STATE Randomly sample one task $\hat{T}_{i}$ from $\mathcal{T}_{up}$
             \STATE Perform inner-loop update of MAML as $\phi_i=\theta_0-\alpha\nabla_{\theta_0}\mathcal{L}(f_{\theta_0},\hat{D}^{s}_{i})$
        \ENDFOR
            \STATE Calculate $\mathcal{L}(f_{\phi_{i}},{D}_{i}^{q})$ and $\mathcal{L}_{adv}(\theta_0, D_i^s, D_i^q)$
            \STATE Update the meta model parameter $\theta_0$ as $\theta_0 \leftarrow \theta_0 - \beta\frac{1}{B}\sum_{i=1}^{n}\nabla_{\theta_0} \mathcal{L}(f_{\phi_{i}},\hat{D}_{i}^{q})$
            \STATE Calculate the objective function in Eq.~(5) and perform backpropagation to update the Task Up-sampling Network
        \ENDWHILE
    \end{algorithmic}
\end{algorithm}
\vspace{5mm}

\section{%The Structures of the Task Up-sampling Network and Training Details.
Network Architecture of the Task Up-sampling Network}
\begin{figure}[h]
    \centering
    \includegraphics[width=4.2in, height = 1.5in]{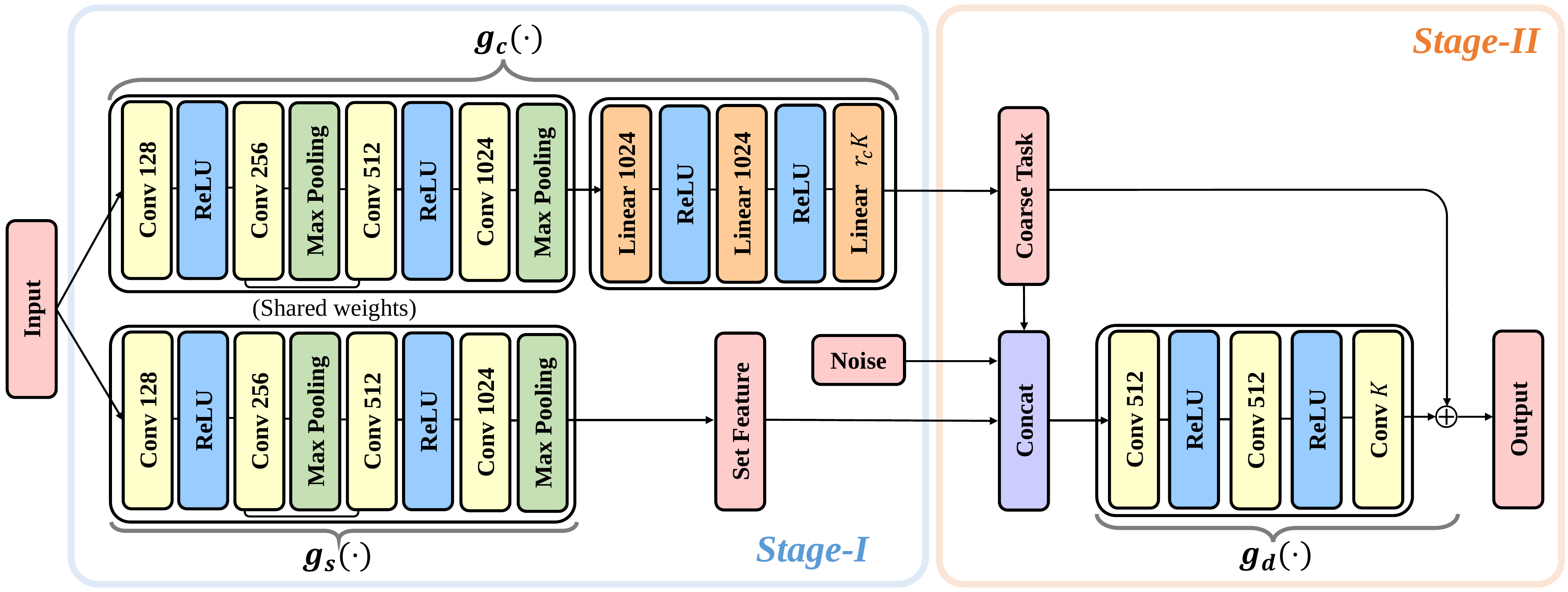}
    \caption{The up-sampling network of the regression task.}
    \label{regression_network}
\end{figure}

In this section, we provide the network architectures for the Task Up-sampling Network for both regression and classification tasks. As shown in Fig.~\ref{regression_network}, the set encoder $g_s(\cdot)$ of the Task Up-Sampling Network for regression tasks consists of 2 convolution layers followed by a max-pooling layer to extract the permutation-invariant feature for the input task patch. The dimension of the set feature is $1024$.
The coarse task generator $g_c(\cdot)$ consists of a set encoder to extract the set feature for the input patch, followed by 3 linear layers to generate coarse tasks from the set feature. The set encoder in coarse generator is the same as $g_s(\cdot)$. The output of the last layers is reshaped into $r_c N_p$ coarse tasks. By concatenating the coarse task and a $r_d$-dimension noise vector, we obtain the input of the decoder $g_d(\cdot)$. 
The decoder consists of 3 convolution layers. We then use the output of the last layer as residual added to the coarse tasks to obtain the up-sampling tasks.

 % (4.2,1.3)
\begin{figure}[h]
    \centering
    \includegraphics[width=4.2in, height = 1.5in]{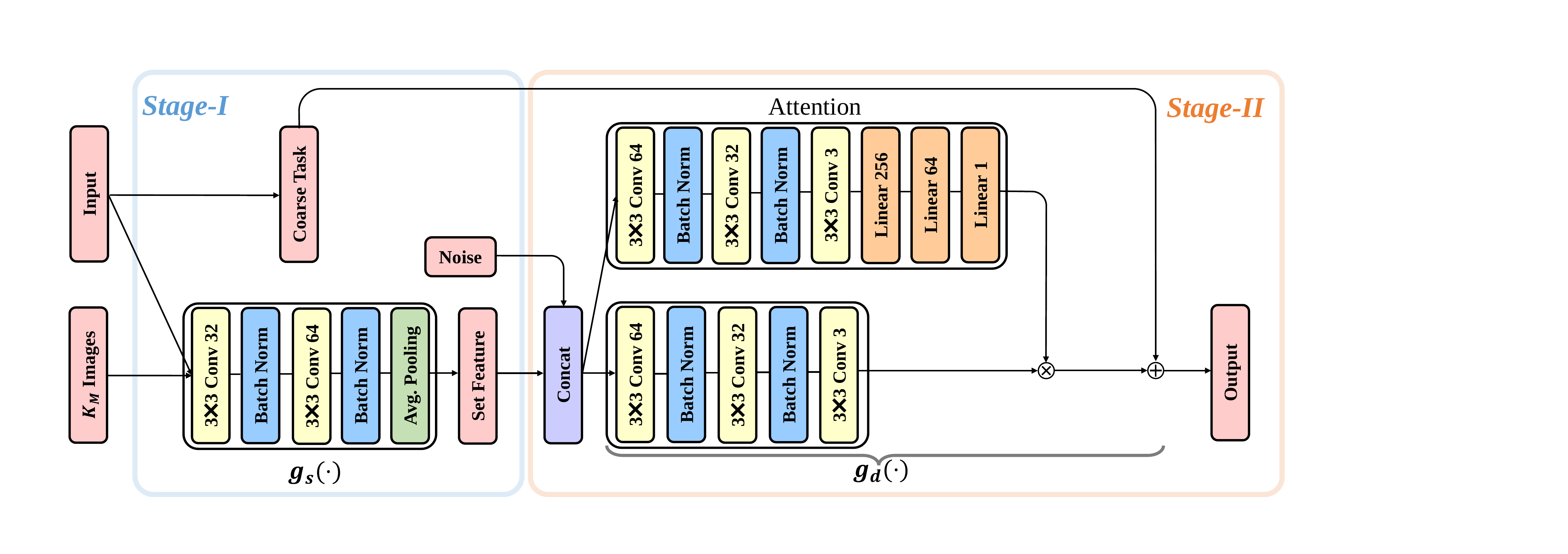}
    \caption{The up-sampling network of the classification task.}
    \label{classification_network}
\end{figure}

% \vspace{2mm}
The network structure of Task Up-sampling Network for classification tasks is presented in Fig.~\ref{classification_network}. We directly use the input task patch as the coarse tasks and, therefore, the coarse task generator $g_c(\cdot)$ is an identity function. The set encoder $g_s(\cdot)$ consists of 2 convolution layers, each followed by a Batch Norm layer. We use the $K_M$ images in the memory bank as the perturbation, concatenating with a $(N_p\!\times\!K_M)$-dimension noise vector to obtain the input to the decoder. The decoder consists of a attention module and a mapping module. The attention module is constructed by 3 convolution layers, followed by 3 linear layers. The attention block generates the attention scores. The mapping module, which consists of 3 convolution layers with xxx filters, maps the $K_M$ perturbation to $K_M$ residual features. We perform weighted sum of the $K_M$ residual features with the attention scores to generate $r$ final residual features and add them to the coarse tasks to obtain the up-sampling 1-shot tasks. We then construct $r$ augmented tasks by stacking $K^s + K^q$ 1-shot tasks.

\section{Setups and Additional Experiment Results for Regression Tasks}

\subsection{Setups of Hyperparameter}
The hyperparameters of the ATU are listed in Table~\ref{reg_hyperparameters}.

\subsection{Effect of Augmentation Ratio}

In the regression task, we assume the combination of the augmented and original tasks will better approximate the real task distribution and, therefore, update the meta-model not only with the augmented tasks generated by the Task Up-sampled Network, but also with the original meta-training tasks. We define the augmentation ratio as the proportion of augmented tasks among all the tasks. Note that the experiment results shown in Table~2 are obtained by setting the augmented ratio as $0.2$. We present the results with other ratios in Table~\ref{reg_aug_ratio}. It can be observed that the performance with positive ratio is better than the performance with ratio$=0$, which means the meta-model is trained without augmented tasks. These results indicates that the augmented tasks are more informative than the original tasks in training a better meta-model.

\begin{minipage}[t!]{\textwidth}
    \begin{minipage}[b]{.45\linewidth}
        \vspace{0pt}
        \centering
        \makeatletter\def\@captype{table}
        \caption{Ablation study on the augmentation ratio of the sine regression task.}
    	\label{reg_aug_ratio}
        \resizebox{65mm}{13mm}{
     	    \begin{tabular}{cccc|cccc} 
    			\toprule[1pt]
    			\multicolumn{4}{l}{\multirow{2}{*}{Augmentation ratio}} & 
    			\multicolumn{4}{l}{\multirow{2}{*}{TU performance (10-shot)}} \\
     			\multicolumn{4}{l}{\multirow{2}{*}{ }} & 
    			\multicolumn{4}{l}{\multirow{2}{*}{ }} 
    		\\   \hline

    			\multicolumn{4}{c}{\multirow{1}{*}{0}} &
    			\multicolumn{4}{c}{\multirow{1}{*}{$0.93\pm 0.18$}} 
    		\\
    
    			\multicolumn{4}{c}{\multirow{1}{*}{0.2}} &
    			\multicolumn{4}{c}{\multirow{1}{*}{$0.84\pm 0.16$}} 
    		\\
    			
    			\multicolumn{4}{c}{\multirow{1}{*}{0.4}} &
    			\multicolumn{4}{c}{\multirow{1}{*}{$0.89\pm 0.17$}} 
    		\\	
    		
    			\multicolumn{4}{c}{\multirow{1}{*}{0.6}} &
    			\multicolumn{4}{c}{\multirow{1}{*}{$0.91\pm 0.18$}} 
    		\\	
    			\bottomrule[1pt]
    		\end{tabular} 
    	}
    \end{minipage}
    \hspace{0.10in}
    \begin{minipage}[b]{.45\linewidth}
        \vspace{0pt}
        \centering
        \makeatletter\def\@captype{table}
		\caption{Hyperparameters of the sine regression task in Table~2.}
    	\label{reg_hyperparameters}
        \resizebox{65mm}{13mm}{
     	    \begin{tabular}{cccc|cccc} 
    			\toprule[1pt]
    			\multicolumn{4}{c}{\multirow{2}{*}{Hyperparameters}} & 
    			\multicolumn{4}{c}{\multirow{2}{*}{ATU}} \\
    			
     			\multicolumn{4}{l}{\multirow{2}{*}{ }} & 
    			\multicolumn{4}{l}{\multirow{2}{*}{ }}
    		\\   \hline
    			
    			\multicolumn{4}{c}{\multirow{1}{*}{maximum training iterations}} &
    			\multicolumn{4}{c}{\multirow{1}{*}{3750}} 
    		\\	
    
    			\multicolumn{4}{c}{\multirow{1}{*}{up-sampling ratio r ($r_c,r_d$)}} &
    			\multicolumn{4}{c}{\multirow{1}{*}{8 (2, 4)}} 
    		\\	
    			
    			\multicolumn{4}{c}{\multirow{1}{*}{loss weights $(\eta_1,\eta_2,\eta_3)$}} &
    			\multicolumn{4}{c}{\multirow{1}{*}{($8e^{-3},4e^{-3},3e^{-1}$)}} 
    		\\
    	
     			\multicolumn{4}{c}{\multirow{1}{*}{size of $\mathcal{T}_g$}} &
    			\multicolumn{4}{c}{\multirow{1}{*}{64}} 
    		\\
    			\bottomrule[1pt]
    		\end{tabular}    
    		}
    \end{minipage}
\end{minipage}

\begin{figure}[h]
\centering
\begin{minipage}[b]{0.5\linewidth}
\subsection{Effect of $d_{EMD}(D_i^s, D_i^q)$}
As presented in Section 4.1, we propose to apply an extra EMD loss $d_{EMD}(D_i^s, D_i^q)$ on the support and query set for each generated task to encourage the points in the generated support set and the query set to follow the same sine curve. In Fig.~3, we have visualized the tasks generated by the Task Up-sampling Network trained with the extra EMD loss. Here we provide additional visualization result for tasks generated by the Task Up-sampling Network trained without the extra EMD loss in Fig.~\ref{regression_without_loss}. It can be seen that the support set and the query set are not on the same sinusoid, indicating the task generated with the adversarial loss may be too difficult or not even a valid task.
\end{minipage}
\hfill
\begin{minipage}[b]{0.45\linewidth}

\includegraphics[width=2.7in, height =1.8in]{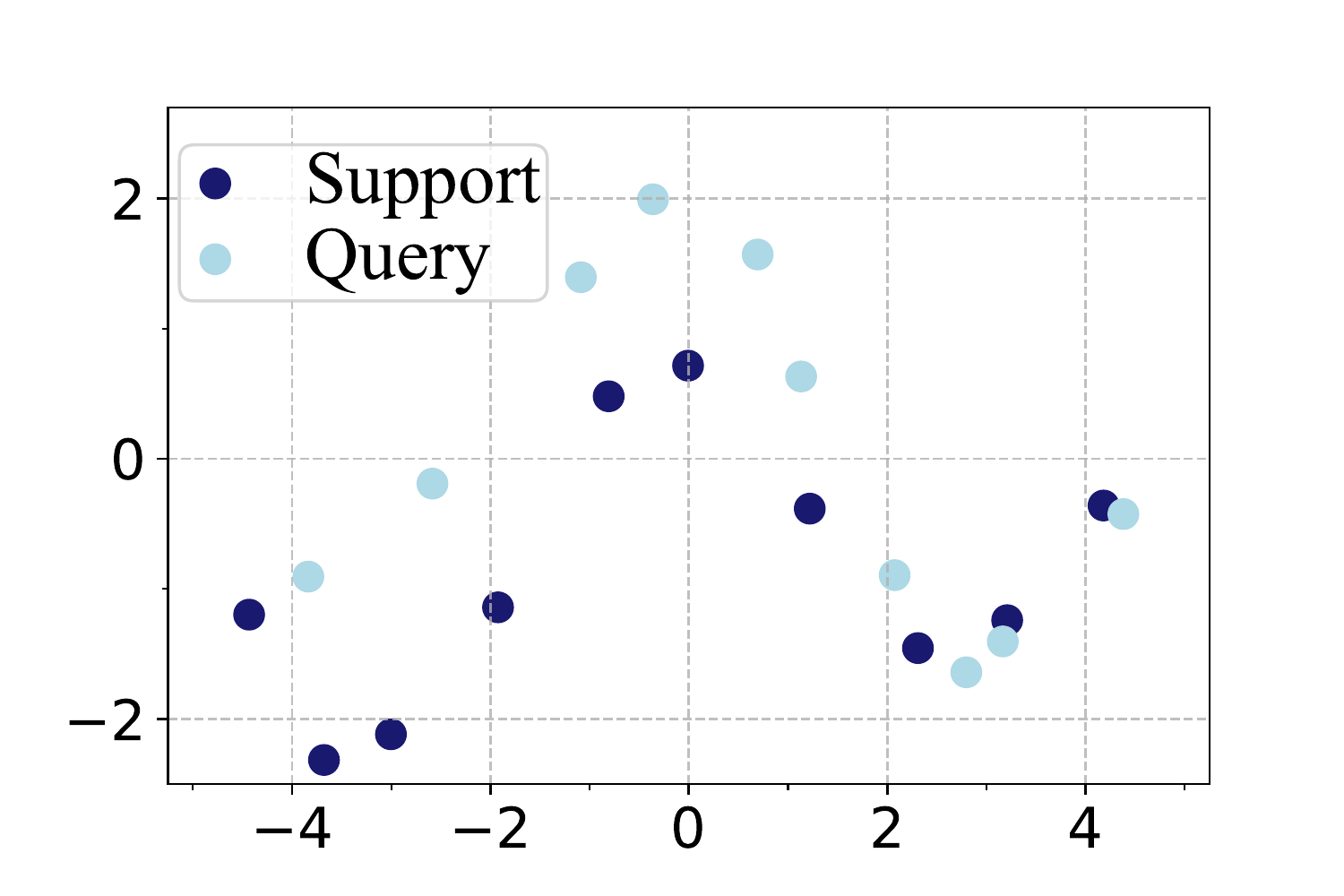}
\caption{The generated tasks generated by ATU when $\eta_3=0$.}
\label{regression_without_loss}
\end{minipage}
\end{figure}

\section{Setups and Additional Experiment Results %in 
for Classification Tasks %and Setup of the Classification Part.
}
\subsection{Introduction and Hyperparameters of the Four Datasets%and Detailed Hyperparameters
}

We provide the detailed information of the datasets and hyperparameters of the classification tasks for obtaining the results in Tabel~4 in this section. We construct the 4 datasets following the settings in MLTI~\cite{yao2021meta}.

\textbf{miniImagenet-S.} Compared with miniImagenet, miniImagenet-S has fewer meta-training classes so as to limit the task number. The specific meta-training classes of miniImagenet-S include:

\noindent \textit{n03017168, n07697537, n02108915, n02113712, n02120079, n04509417,n02089867, n03888605, 
n04258138, n03347037, n02606052, n06794110}

We use four convolutional blocks and a classifier as the base learner~\cite{yao2021meta,finn2017model}, and each convolutional block contains a convolutional layer, a batch normalization layer and a ReLU activation layer. In order to analyze the effect of the number of meta-training tasks, we add more classes for meta-training according to the following sequence:

\noindent \textit{n03476684, n02966193, n13133613, n03337140, n03220513, n03908618,n01532829, n04067472, 
n02074367, ~n03400231, n02108089, n01910747, n02747177, n02795169, n04389033, n04435653,
n02111277, n02108551,\\n04443257, n02101006, n02823428, n03047690, n04275548, n04604644,
n02091831, n01843383, n02165456, n03676483, n04243546, n03527444, n01770081, n02687172, 
n09246464, n03998194, n02105505, n01749939, n04251144, n07584110, n07747607, n04612504,
n01558993, n03062245, n04296562, n04596742, n03838899, n02457408, n13054560, n03924679,
n03854065, n01704323, n04515003, n03207743}

\textbf{ISIC.} ISIC skin dataset~\cite{milton2019automated} was provided by ISIC2018 Challenge, in which 7 disease classes and 10015 dermoscopic images are included. Following~\cite{yao2021meta,li2020difficulty}, \textit{Nevus, Malanoma, Benign Keratoses, Basal Cell Carcinoma}, the four categories with the largest number of images, are as meta-training classes; the rest \textit{Dermatofibroma, Pigmented Bowen's, Benign Keratoses} are as meta-testing classes. We re-scale the size of each medical image to $84\times84\times3$ and adopt the same 4-layer convolutional as the base model like miniImagenet-S.

\textbf{DermNet-S.} Dermnet-S are part of the public Dermnet Skin Disease Altas, in which 625 different fine-grained categories are included. Dermnet-S chooses the top-30 classes for meta-training\kai{. T}%,t
he concrete meta-training classes and meta-testing classes are:
\begin{itemize}
    \item \textbf{Meta-training classes:} \textit{Seborrheic Keratoses Ruff, Herpes Zoster, Atopic Dermatitis Adult Phase, Psoriasis Chronic Plaque, Eczema Hand, Seborrheic Dermatitis, Keratoacanthoma, Lichen Planus, Epidermal Cyst, Eczema Nummular, Tinea (Ringworm) Versicolor, Tinea (Ringworm) Body, Lichen Simplex Chronicus, Scabies, Psoriasis Palms Soles, Malignant Melanoma, Candidiasis large Skin Folds, Pityriasis Rosea, Granuloma Annulare, Erythema Multiforme, Seborrheic Keratosis Irritated, Stasis Dermatitis and Ulcers, Distal Subungual Onychomycosis, Allergic Contact Dermatitis, Psoriasis, Molluscum Contagiosum, Acne Cystic, Perioral Dermatitis, Vasculitis, Eczema Fingertips.}
    \item \textbf{Meta-testing classes:} \textit{Warts, Ichthyosis Sex Linked, Atypical Nevi, Venous Lake, Erythema Nodosum, Granulation Tissue, Basal Cell Carcinoma Face, Acne Closed Comedo, Scleroderma, Crest Syndrome, Ichthyosis Other Forms, Psoriasis Inversus, Kaposi Sarcoma, Trauma, Polymorphous Light Eruption, Dermagraphism, Lichen Sclerosis Vulva, Pseudomonas, Cutaneous Larva Migrans, Psoriasis Nails, Corns, Lichen Sclerosus Penis, Staphylococcal Folliculitis, Chilblains Perniosis, Psoriasis Erythrodermic, Squamous Cell Carcinoma Ear, Basal Cell Carcinoma Ear, Ichthyosis Dominant, Erythema Infectiosum, Actinic Keratosis Hand, Basal Cell Carcinoma Lid, Amyloidosis, Spiders, Erosio Interdigitalis Blastomycetica, Scarlet Fever, Pompholyx, Melasma, Eczema Trunk Generalized, Metastasis, Warts Cryotherapy, Nevus Spilus, Basal Cell Carcinoma Lip, Enterovirus, Pseudomonas Cellulitis, Benign Familial Chronic Pemphigus, Pressure Urticaria, Halo Nevus, Pityriasis Alba, Pemphigus Foliaceous, Cherry Angioma, Chapped Fissured Feet, Herpes Buttocks, Ridging Beading.}
\end{itemize}

\textbf{Tabular Murris.} The Tabular Murris is a gene dataset (i.e., 2866-dim features) including 105,960 cells of 124 cell types extracted from 23 organs. Following~\cite{yao2021meta,cao2020concept}, the concrete training/validation/testing split is:

\begin{itemize}
    \item \textbf{Meta-training classes:} \textit{BAT, MAT,Limb Muscle, Trachea, Heart, Spleen, GAT, SCAT, Mammary Gland, Liver, Kidney, Bladder, Brain Myeloid, Brain Non-Myeloid, Diaphragm.}
    \item \textbf{Meta-validation classes:} \textit{Skin, Lung, Thymus, Aorta}
    \item \textbf{Meta-testing organs:} \textit{Large Intestine, Marrow, Pancreas, Tongue}
\end{itemize}

Unlike the base model for the other 3 datasets, we use two fully connected blocks and a linear layer as the backbone network, where each fully connected block includes a linear layer, a batch normalization layer, a ReLu activation layer, and a dropout layer. The dropout ratio and the feature channels of the linear layer are set 0.2, 64, which is the same as the settings of~\cite{yao2021meta,cao2020concept}.

For a fair comparison with MLTI, we adopt the same MetaMix strategy as MLTI to augment the query set for the four datasets. We set the augmentation ratio as 1 in classification tasks and do not use the original sampled tasks in meta-training because we empirically find that ATU obtains better performance with higher augmentation ratio.
The other settings are the same with MLTI. More details about the hyperparameters are listed in Table~\ref{Hyper-classification}.
\begin{table*}[t]
    \centering
	\caption{Hyperparameters of Tabel~4}
	\label{Hyper-classification}
 	\setlength{\tabcolsep}{1.8 mm}{
 	    \begin{tabular}{cccccccc|cccc|cccc|cccc|cccc} 
			\toprule[1pt]
			\multicolumn{8}{l|}{Hyperparameters(ATU)} &
			\multicolumn{4}{l|}{miniImagenet-S} &
			\multicolumn{4}{l|}{ISIC} &
			\multicolumn{4}{l|}{Dermnet-S} &
			\multicolumn{4}{l}{Tabular Murris} \\
			
			 \midrule
			\multicolumn{8}{l|}{inner-loop learning rate}&
			\multicolumn{4}{c|}{0.01}&
			\multicolumn{4}{c|}{0.01} &
			\multicolumn{4}{c|}{0.01} &
			\multicolumn{4}{c}{0.01}\\		
			
			\multicolumn{8}{l|}{outer-loop learning rate}&
			\multicolumn{4}{c|}{0.001}&
			\multicolumn{4}{c|}{0.001} &
			\multicolumn{4}{c|}{0.001} &
			\multicolumn{4}{c}{0.001}\\			

			\multicolumn{8}{l|}{$Beta(\delta_1,\delta_2)$}&
			\multicolumn{4}{c|}{(3,5)}&
			\multicolumn{4}{c|}{(2,2)} &
			\multicolumn{4}{c|}{(2,2)} &
			\multicolumn{4}{c}{(2,2)}\\	
			
			\multicolumn{8}{l|}{%num updates
			Number of steps in inner loop}&
			\multicolumn{4}{c|}{5}&
			\multicolumn{4}{c|}{5} &
			\multicolumn{4}{c|}{5} &
			\multicolumn{4}{c}{5}\\				

			\multicolumn{8}{l|}{batch size}&
			\multicolumn{4}{c|}{4}&
			\multicolumn{4}{c|}{4} &
			\multicolumn{4}{c|}{4} &
			\multicolumn{4}{c}{4}\\	

			\multicolumn{8}{l|}{query size %for
			in meta-training tasks}&
			\multicolumn{4}{c|}{15}&
			\multicolumn{4}{c|}{15} &
			\multicolumn{4}{c|}{15} &
			\multicolumn{4}{c}{15}\\				

			\multicolumn{8}{l|}{maximum training iterations}&
			\multicolumn{4}{c|}{50,000}&
			\multicolumn{4}{c|}{50,000} &
			\multicolumn{4}{c|}{50,000} &
			\multicolumn{4}{c}{50,000}\\			

			\multicolumn{8}{l|}{adversarial loss weights $\eta$ ($\eta_1=\eta_2$)}&
			\multicolumn{4}{c|}{3}&
			\multicolumn{4}{c|}{3} &
			\multicolumn{4}{c|}{0.5} &
			\multicolumn{4}{c}{0.5}\\	
			
			\multicolumn{8}{l|}{up-sampling ratio r}&
			\multicolumn{4}{c|}{2}&
			\multicolumn{4}{c|}{2} &
			\multicolumn{4}{c|}{2} &
			\multicolumn{4}{c}{2}\\		
			\bottomrule[1pt]
		\end{tabular}
		}   
\end{table*}

\begin{minipage}[t]{\textwidth}
    \begin{minipage}[b]{.45\linewidth}
        \vspace{0pt}
        \centering
        \makeatletter\def\@captype{table}
        \caption{Ablation study on the memory bank size in the classification task.}
    	\label{K_M-size}
	\setlength{\tabcolsep}{1.5 mm}{
     	    \begin{tabular}{cccc|cccc} 
    			\toprule[1pt]
    			\multicolumn{4}{l}{\multirow{2}{*}{$K_M$}} & 
    			\multicolumn{4}{l}{\multirow{2}{*}{TU(miniImagenet-S 1-shot)}} \\
     			\multicolumn{4}{l}{\multirow{2}{*}{ }} & 
    			\multicolumn{4}{l}{\multirow{2}{*}{ }} 
    		\\   \hline
    
    			\multicolumn{4}{c}{\multirow{1}{*}{3}} &
    			\multicolumn{4}{c}{\multirow{1}{*}{42.16 $\pm$ 0.73\%}} 
    		\\
    
    			\multicolumn{4}{c}{\multirow{1}{*}{5}} &
    			\multicolumn{4}{c}{\multirow{1}{*}{42.20 $\pm$ 0.76\%}} 
    		\\
    			
    			\multicolumn{4}{c}{\multirow{1}{*}{7}} &
    			\multicolumn{4}{c}{\multirow{1}{*}{42.28 $\pm$ 0.72\%}} 
    		\\
    			\bottomrule[1pt]
    		\end{tabular} 
    	}
    \end{minipage}
    \hspace{0.10in}
    \begin{minipage}[b]{.45\linewidth}
        \vspace{0pt}
        \centering
        \makeatletter\def\@captype{table}
		\caption{Sensitivity analysis of the adversarial loss weights on the classification task.}
    	\label{cla_adv_ablation}
    	 	\scalebox{1}{
     	    \begin{tabular}{cccc|cccc} 
    			\toprule[1pt]
    			\multicolumn{4}{c}{\multirow{2}{*}{Adversarial weights}} & 
    			\multicolumn{4}{c}{\multirow{2}{*}{ATU (mini-S 1-shot)}} \\
     			\multicolumn{4}{l}{\multirow{2}{*}{ }} & 
    			\multicolumn{4}{l}{\multirow{2}{*}{ }}
            \\ \hline
    
    			\multicolumn{4}{c}{\multirow{1}{*}{$\eta_1=\eta_2=1$}} &
    			\multicolumn{4}{c}{\multirow{1}{*}{42.38 $\pm$ 0.82\%}} 
    		\\	
    
    			\multicolumn{4}{c}{\multirow{1}{*}{$\eta_1=\eta_2=3$}} &
    			\multicolumn{4}{c}{\multirow{1}{*}{42.60 $\pm$ 0.84\%}} 
    		\\	
    			
    			\multicolumn{4}{c}{\multirow{1}{*}{$\eta_1=\eta_2=5$}} &
    			\multicolumn{4}{c}{\multirow{1}{*}{41.67 $\pm$ 0.79\%}}
    		\\
    			\bottomrule[1pt]
    		\end{tabular}    
    		}
    \end{minipage}
\end{minipage}

\subsection{Ablation Study% and Overfitting Analysis
}

\textbf{Effect of $K_M$.} As shown in Table~\ref{K_M-size}, the classification performance increases as the memory bank size $K_M$ increases. But the performance gain is not very significant for a large $K_M$. Considering the training efficiency, we set $K_M=3$ in all classification experiments.
\vspace{-3pt}
\begin{table}[h]
    \vspace{-2pt}
    \centering
    \makeatletter\def\@captype{table}
	\caption{The averaged accuracy with 95$\%$ confidence intervals of various interpolation task augmentation methods and our task up-sampling method on miniImagenet-s (5-shot).}
	\label{MB_baseline}
	\setlength{\tabcolsep}{2 mm}{
 	    \begin{tabular}{cccc|cccc} 
			\toprule[1pt]
			\multicolumn{4}{l}{\multirow{2}{*}{Task generation method}} & 
			\multicolumn{4}{l}{\multirow{2}{*}{miniImagenet-S (5-shot)}} \\
 
 			\multicolumn{4}{l}{\multirow{2}{*}{ }} & 
			\multicolumn{4}{c}{\multirow{2}{*}{ }} \\
			\midrule
			
			\multicolumn{4}{c}{\multirow{1}{*}{~ Naive Baseline$^{1}$}} &
			\multicolumn{4}{c}{\multirow{1}{*}{$53.49\pm 0.74\%$}} \\	
			
			\multicolumn{4}{c}{\multirow{1}{*}{~ Naive Baseline$^{2}$}} &
			\multicolumn{4}{c}{\multirow{1}{*}{$50.25\pm 0.71\%$}} \\	
			
			\multicolumn{4}{c}{\multirow{1}{*}{~ Naive Baseline$^{3}$}} &
			\multicolumn{4}{c}{\multirow{1}{*}{$53.91\pm 0.78\%$}} \\	
			
			\multicolumn{4}{c}{\multirow{1}{*}{TU}} &
			\multicolumn{4}{c}{\multirow{1}{*}{$56.33\pm 0.79\%$}} \\	
			\bottomrule[1pt]
		\end{tabular}   
		}
\end{table}

\textbf{Effect of $\eta_1$,$\eta_2$.} We %also 
choose different ($\eta_1$,$\eta_2$) %in Table~\ref{cla_adv_ablation} 
to explore the sensitivity of model performance to adversarial loss weights. It can be observed from the results in Table~\ref{cla_adv_ablation} that the adversarial loss weights have a large influence on the performance of the model and it achieves the best performance when setting $\eta_1=\eta_2=3$.

\textbf{Effect of Augmented Task Generation Strategies.}
Due to the high complexity of the classification tasks' distribution, we assume its latent task distribution is smooth and construct the ground-truth task manifold $\mathcal{T}_g$ via mixing all image in each %whole 
class of $T_i$ with its corresponding nearest image in the memory bank (the sampled $K_M$ images) (see Algorithm~\ref{alg:classification_meta_train}). Under this assumption, one naive method is to generate augmented tasks in the same way as the generation of ground-truth tasks where we can %is just to 
directly mix the images in the tasks with the $K_M$ images in the memory bank. To verify the effectiveness and necessity of training a Task Up-sampling Network, we compare TU with 3 naive methods in Table~\ref{MB_baseline}: (1) Naive Baseline$^{1}$:%, 
for each image of task $T_{i}$, we randomly choose one image in the memory bank to mix; (2) Naive Baseline$^{2}$: for all images in a class of task $T_{i}$, we randomly choose one image in the memory bank to mix; (3) Naive Baseline$^{3}$: for all images in a class of task $T_{i}$, we choose the nearest image in the memory bank to mix. The Naive Baseline$^{3}$ is the
method that we used to construct $\mathcal{T}_{g}$. The results in Table \ref{MB_baseline} show that TU outperforms the other 3 baselines by a large margin.
TU outperforms Naive Baseline$^{3}$ because the tasks generated by TU match the local task distribution better than those generated by just mixing with the images in the memory bank. Moreover, the tasks generated by TU is more diverse and informative. Taking this into consideration, we set the augmentation ratio to be 1 and do not use the original tasks in the meta-training.

\subsection{Visualization of the generated Classification tasks of $\mathcal{T}_{up}$.}
\begin{figure}[h]
    \centering
    \includegraphics[width=4.1in, height = 2.5in]{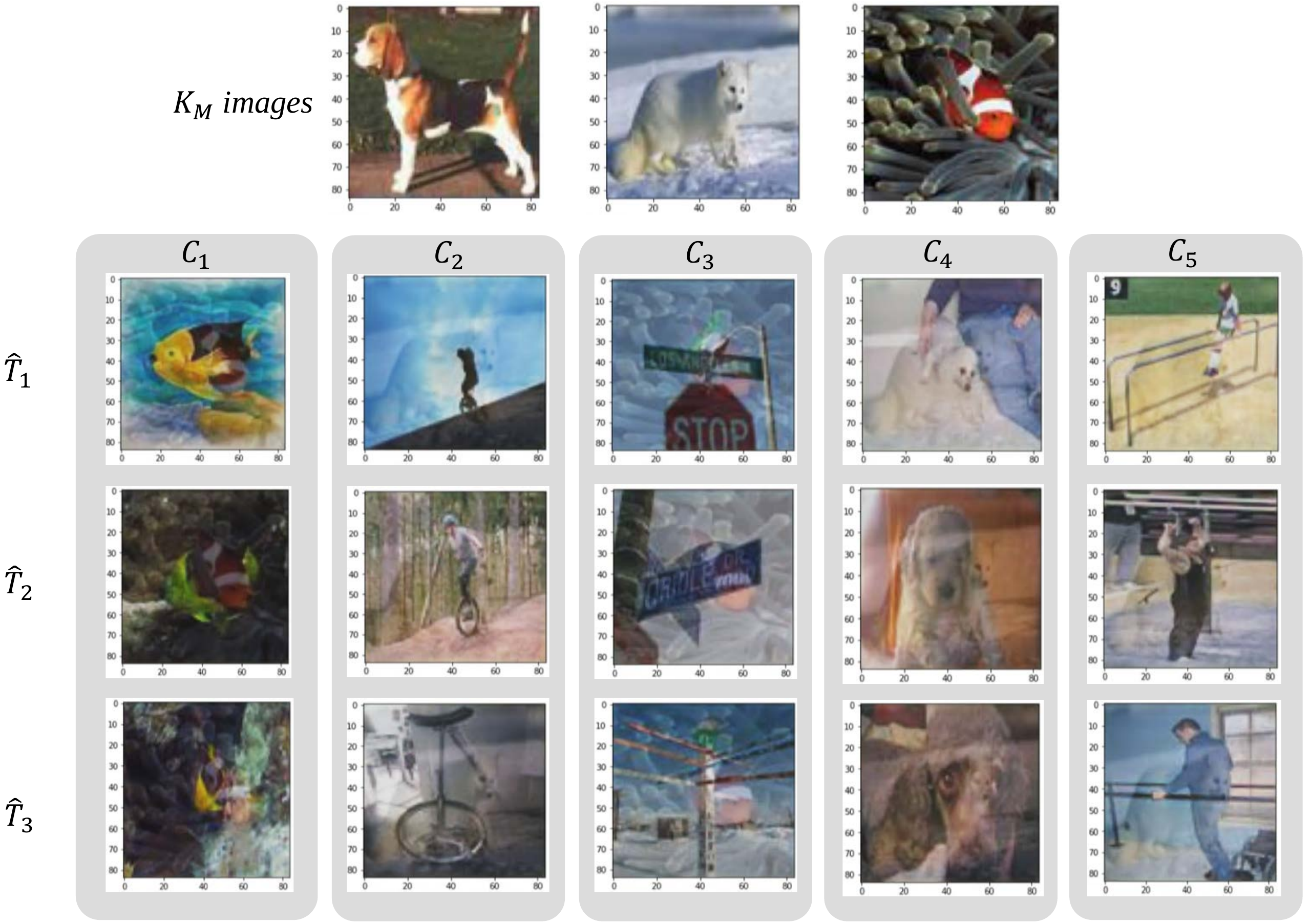}
    \caption{Visualization of part of the up-sampled classification tasks (i.e., $\mathcal{T}_{up}$) generated by ATU.}
    \label{vis_classification_tasks}
\end{figure}

We visualize part of the images in a generated classification task in  Fig.~\ref{vis_classification_tasks}. The three images in the top row are the sampled extra $K_M$ images and $\hat{T}_1, \hat{T}_2, \hat{T}_3$ are three 5-way 1-shot tasks generated by ATU.

\subsection{Complete Results with Confidence Interval}
We list the complete results with 95\% confidence interval in Table~\ref{full_classification},\ref{full_backbones},\ref{full_cross},  which are corresponding to the Tabel~4,~5,~6 in Section 6.2.

\vspace{-5pt}
\begin{table}[h]
\centering
\caption{Complete results of Table~5 with 95\% confidence interval under different backbones.}
\label{full_backbones}
 	    \begin{tabular}{ccc|cc|cc|cc|cc} 
		%\begin{tabular}{ccc|ccc|ccc} 
			\toprule
			\multicolumn{3}{l|}{\multirow{1}{*}{Method}} &
% 			\multicolumn{2}{c|}{\multirow{1}{*}{Strategies}} & 
			\multicolumn{2}{c|}{\multirow{1}{*}{miniImagenet-S}} & 
			\multicolumn{2}{c|}{\multirow{1}{*}{ISIC}} & 
			\multicolumn{2}{c|}{\multirow{1}{*}{DermNet-S}} & 
			\multicolumn{2}{c}{\multirow{1}{*}{Tabular Muris}} \\
			
			\midrule
		    %\multicolumn{1}{c}{\multirow{1}{*}{\qquad\textit{MetaSGD as backbone}}}   \\ 
			\multicolumn{3}{l|}{\multirow{1}{*}{MetaSGD~\cite{li2017meta}}} &
% 			\multicolumn{2}{c|}{\multirow{1}{*}{}} & 
			\multicolumn{2}{c|}{\multirow{1}{*}{$37.88\pm 0.74\%$}} & 
			\multicolumn{2}{c|}{\multirow{1}{*}{$58.79\pm 0.82\%$}} &
			\multicolumn{2}{c|}{\multirow{1}{*}{$42.07\pm 0.83\%$}} & 
			\multicolumn{2}{c}{\multirow{1}{*}{$81.55 \pm 0.91\%$}}  \\

			\multicolumn{3}{l|}{\multirow{1}{*}{MetaSGD+MLTI}} &
% 			\multicolumn{2}{c|}{\multirow{1}{*}{+MLTI}} & 
			\multicolumn{2}{c|}{\multirow{1}{*}{$39.58\pm 0.76\%$}} & 
			\multicolumn{2}{c|}{\multirow{1}{*}{$61.57\pm 1.10\%$}} & 
			\multicolumn{2}{c|}{\multirow{1}{*}{$45.49\pm 0.83\%$}} & 
			\multicolumn{2}{c}{\multirow{1}{*}{$83.31\pm 0.87\%$}}  \\

			\multicolumn{3}{l|}{\multirow{1}{*}{MetaSGD+ATU}} &
% 			\multicolumn{2}{c|}{\multirow{1}{*}{+ATU}} & 
			\multicolumn{2}{c|}{\multirow{1}{*}{$\textbf{40.52}\pm\textbf{0.78\%}$}} & 
			\multicolumn{2}{c|}{\multirow{1}{*}{$\textbf{62.84}\pm\textbf{1.01\%}$}} & 
			\multicolumn{2}{c|}{\multirow{1}{*}{$\textbf{46.78}\pm\textbf{0.84\%}$}} & 
			\multicolumn{2}{c}{\multirow{1}{*}{$\textbf{83.84}\pm\textbf{0.90\%}$}}  \\
            
			\midrule
		    %\multicolumn{1}{c}{\multirow{1}{*}{\qquad\textit{ANIL as backbone}}}   \\ 			
			
			\multicolumn{3}{l|}{\multirow{1}{*}{ANIL~\cite{raghu2019rapid}}} &
% 			\multicolumn{2}{c|}{\multirow{1}{*}{}} & 
			\multicolumn{2}{c|}{\multirow{1}{*}{$38.02\pm 0.75\%$}} & 
			\multicolumn{2}{c|}{\multirow{1}{*}{$59.48\pm 1.00\%$}} &
			\multicolumn{2}{c|}{\multirow{1}{*}{$44.58\pm 0.85\%$}} & 
			\multicolumn{2}{c}{\multirow{1}{*}{$75.67\pm 0.99\%$}}  \\

			\multicolumn{3}{l|}{\multirow{1}{*}{ANIL+MLTI}} &
% 			\multicolumn{2}{c|}{\multirow{1}{*}{+MLTI}} & 
			\multicolumn{2}{c|}{\multirow{1}{*}{$39.15\pm0.73\%$}} & 
			\multicolumn{2}{c|}{\multirow{1}{*}{$61.78\pm1.24\%$}} & 
			\multicolumn{2}{c|}{\multirow{1}{*}{$46.79\pm0.77\%$}} & 
			\multicolumn{2}{c}{\multirow{1}{*}{$77.11\pm1.00\%$}}  \\

			\multicolumn{3}{l|}{\multirow{1}{*}{ANIL+ATU}} &
% 			\multicolumn{2}{c|}{\multirow{1}{*}{+ATU}} & 
			\multicolumn{2}{c|}{\multirow{1}{*}{$\textbf{39.27}\pm\textbf{0.76\%}$}} & 
			\multicolumn{2}{c|}{\multirow{1}{*}{$\textbf{62.12}\pm\textbf{0.98\%}$}} & 
			\multicolumn{2}{c|}{\multirow{1}{*}{$\textbf{47.03}\pm\textbf{0.85\%}$}} & 
			\multicolumn{2}{c}{\multirow{1}{*}{$\textbf{77.23}\pm\textbf{0.99\%}$}}  \\	
			\bottomrule
		\end{tabular}
		\end{table}

\vspace{-10pt}
\begin{table}[h]
\centering
\caption{Complete results of Table~6 with 95\% confidence interval under the cross-domain setting.}
\label{full_cross}
 	    \begin{tabular}{ccc|cccc|cccc} 
			\toprule
			\multicolumn{3}{l|}{\multirow{2}{*}{Model}} &
			\multicolumn{4}{c|}{\multirow{1}{*}{miniImagenet-S$\rightarrow$ DermNet-S}} &
			\multicolumn{4}{c}{\multirow{1}{*}{DermNet-S$\rightarrow$ miniImagenet-S}} \\
			\multicolumn{3}{l|}{\multirow{1}{*}{}} & 
% 			\multicolumn{2}{c|}{\multirow{1}{*}{}} &
			\multicolumn{2}{c}{\multirow{1}{*}{1-shot}} &
			\multicolumn{2}{c|}{\multirow{1}{*}{5-shot}} &
			\multicolumn{2}{c}{\multirow{1}{*}{1-shot}} &
			\multicolumn{2}{c}{\multirow{1}{*}{5-shot}}\\	
            
            \midrule
		    %\multicolumn{1}{c}{\multirow{1}{*}{\qquad %\textit{MAML as backbone}}}   \\ 
			\multicolumn{3}{l|}{\multirow{1}{*}{MAML~\cite{finn2017model}}} & 
% 			\multicolumn{2}{l|}{\multirow{1}{*}{}} &
			\multicolumn{2}{c}{\multirow{1}{*}{$34.46 \pm 0.63\%$}} &
			\multicolumn{2}{c|}{\multirow{1}{*}{$50.36 \pm 0.64\%$}} &
			\multicolumn{2}{c}{\multirow{1}{*}{$28.78 \pm 0.55\%$}} &
			\multicolumn{2}{c}{\multirow{1}{*}{$41.29 \pm 0.64\%$}}  \\

			\multicolumn{3}{l|}{\multirow{1}{*}{MAML+ATU}} & 
% 			\multicolumn{2}{l|}{\multirow{1}{*}{+ATU}} &
			\multicolumn{2}{c}{\multirow{1}{*}{$\textbf{36.86}\pm \textbf{0.64\%}$}} &
			\multicolumn{2}{c|}{\multirow{1}{*}{$\textbf{51.98}\pm \textbf{0.62\%}$}} &
			\multicolumn{2}{c}{\multirow{1}{*}{$\textbf{30.68}\pm \textbf{0.68\%}$}} &
			\multicolumn{2}{c}{\multirow{1}{*}{$\textbf{46.72}\pm \textbf{0.73\%}$}}  \\		    

            \midrule
		    %\multicolumn{1}{c}{\multirow{1}{*}{\qquad \textit{MetaSGD as backbone}}}   \\ 
			\multicolumn{3}{l|}{\multirow{1}{*}{MetaSGD~\cite{li2017meta}}} & 
% 			\multicolumn{2}{l|}{\multirow{1}{*}{}} &
			\multicolumn{2}{c}{\multirow{1}{*}{$31.07\pm 0.57\%$}} &
			\multicolumn{2}{c|}{\multirow{1}{*}{$49.07\pm 0.59\%$}} &
			\multicolumn{2}{c}{\multirow{1}{*}{$28.17\pm 0.53\%$}} &
			\multicolumn{2}{c}{\multirow{1}{*}{$41.83\pm 0.67\%$}}  \\		
			
			\multicolumn{3}{l|}{\multirow{1}{*}{MetaSGD+ATU}} & 
% 			\multicolumn{2}{l|}{\multirow{1}{*}{ATU}} &
			\multicolumn{2}{c}{\multirow{1}{*}{$\textbf{37.75}\pm\textbf{0.65\%}$}} &
			\multicolumn{2}{c|}{\multirow{1}{*}{$\textbf{54.60}\pm\textbf{0.58\%}$}} &
			\multicolumn{2}{c}{\multirow{1}{*}{$\textbf{30.78}\pm\textbf{0.58\%}$}} &
			\multicolumn{2}{c}{\multirow{1}{*}{$\textbf{44.01}\pm\textbf{0.68\%}$}}  \\	
			\bottomrule
		\end{tabular}
\end{table}

\vspace{-3pt}
\begin{table}[h]
\centering
\caption{Complete classification results of Table~4 with 95\% confidence interval.}
\label{full_classification}
\setlength{\tabcolsep}{1 mm}{
 	    \begin{tabular}{cc|ccc|cccc|cccc|cccc|cccc} 
			\toprule[1pt]
			\multicolumn{2}{l}{\multirow{1}{*}{Setting}} &
			\multicolumn{3}{c|}{\multirow{1}{*}{Model}} & 
			\multicolumn{4}{c|}{\multirow{1}{*}{miniImagenet-S}} &
			\multicolumn{4}{c|}{\multirow{1}{*}{ISIC}} &
			\multicolumn{4}{c|}{\multirow{1}{*}{DermNet-S}} &
			\multicolumn{4}{c}{\multirow{1}{*}{Tabular Murris}}\\
			\midrule

			\multicolumn{2}{l}{\multirow{9}{*}{1-shot}} & 
			\multicolumn{3}{l|}{\multirow{1}{*}{MAML~\cite{finn2017model}} } & 
			\multicolumn{4}{c|}{\multirow{1}{*}{$38.27\pm 0.74\%$}} &
			\multicolumn{4}{c|}{\multirow{1}{*}{$57.59\pm 0.79\%$}} &
			\multicolumn{4}{c|}{\multirow{1}{*}{$43.47\pm 0.83\%$}} &
			\multicolumn{4}{c}{\multirow{1}{*}{$79.08\pm 0.91\%$}} \\	 
			
			\multicolumn{2}{l}{\multirow{9}{*}{}} & 
			\multicolumn{3}{l|}{\multirow{1}{*}{Meta-Reg~\cite{yin2020meta}} } & 
			\multicolumn{4}{c|}{\multirow{1}{*}{$38.35\pm 0.76\%$}} &
			\multicolumn{4}{c|}{\multirow{1}{*}{$58.57\pm 0.94\%$}} &
			\multicolumn{4}{c|}{\multirow{1}{*}{$45.01\pm 0.83\%$}} &
			\multicolumn{4}{c}{\multirow{1}{*}{$79.18\pm 0.87\%$}} \\
			
			\multicolumn{2}{l}{\multirow{9}{*}{}} & 
			\multicolumn{3}{l|}{\multirow{1}{*}{TAML~\cite{jamal2019task}} } & 
			\multicolumn{4}{c|}{\multirow{1}{*}{$38.70\pm 0.77\%$}} &
			\multicolumn{4}{c|}{\multirow{1}{*}{$58.39\pm 1.00\%$}} &
			\multicolumn{4}{c|}{\multirow{1}{*}{$45.73\pm 0.84\%$}} &
			\multicolumn{4}{c}{\multirow{1}{*}{$79.82\pm 0.87\%$}} \\
			
			\multicolumn{2}{l}{\multirow{9}{*}{}} & 
			\multicolumn{3}{l|}{\multirow{1}{*}{Meta-Dropout~\cite{lee2019meta}}} & 
			\multicolumn{4}{c|}{\multirow{1}{*}{$38.32\pm 0.75\%$}} &
			\multicolumn{4}{c|}{\multirow{1}{*}{$58.40\pm 1.02\%$}} &
			\multicolumn{4}{c|}{\multirow{1}{*}{$44.30\pm 0.84\%$}} &
			\multicolumn{4}{c}{\multirow{1}{*}{$78.18\pm 0.93\%$}} \\
			
			\multicolumn{2}{l}{\multirow{9}{*}{}} & 
			\multicolumn{3}{l|}{\multirow{1}{*}{MetaMix~\cite{yao2021improving}}} & 
			\multicolumn{4}{c|}{\multirow{1}{*}{$39.43 \pm 0.77\%$}} &
			\multicolumn{4}{c|}{\multirow{1}{*}{$60.34 \pm 1.03\%$}} &
			\multicolumn{4}{c|}{\multirow{1}{*}{$46.81 \pm 0.81\%$}} &
			\multicolumn{4}{c}{\multirow{1}{*}{$81.06\pm 0.86\%$}} \\
			
			\multicolumn{2}{l}{\multirow{9}{*}{}} & 
			\multicolumn{3}{l|}{\multirow{1}{*}{Meta-Maxup~\cite{ni2021data}} } & 
			\multicolumn{4}{c|}{\multirow{1}{*}{$39.28 \pm 0.77\%$}} &
			\multicolumn{4}{c|}{\multirow{1}{*}{$58.68 \pm 0.86\%$}} &
			\multicolumn{4}{c|}{\multirow{1}{*}{$46.10 \pm 0.82\%$}} &
			\multicolumn{4}{c}{\multirow{1}{*}{$79.56 \pm 0.89\%$}} \\
			
			\multicolumn{2}{l}{\multirow{9}{*}{}} & 
			\multicolumn{3}{l|}{\multirow{1}{*}{MLTI~\cite{yao2021meta}} } & 
			\multicolumn{4}{c|}{\multirow{1}{*}{$41.58 \pm 0.72\%$}} &
			\multicolumn{4}{c|}{\multirow{1}{*}{$61.79 \pm 1.00\%$}} &
			\multicolumn{4}{c|}{\multirow{1}{*}{$48.03 \pm 0.79\%$}} &
			\multicolumn{4}{c}{\multirow{1}{*}{$81.73 \pm 0.89\%$}} \\
			\cline{3-21}
			
			\multicolumn{2}{l}{\multirow{9}{*}{}} & 
			\multicolumn{3}{l|}{\multirow{1}{*}{TU }} & 
			\multicolumn{4}{c|}{\multirow{1}{*}{$42.16\pm 0.76\%$}} &
			\multicolumn{4}{c|}{\multirow{1}{*}{$62.03\pm 0.95\%$}} &
			\multicolumn{4}{c|}{\multirow{1}{*}{$48.07\pm 0.83\%$}} &
			\multicolumn{4}{c}{\multirow{1}{*}{$81.88\pm 0.90\%$}} \\

			\multicolumn{2}{l}{\multirow{9}{*}{}} & 
			\multicolumn{3}{l|}{\multirow{1}{*}{ATU }} & 
			\multicolumn{4}{c|}{\multirow{1}{*}{$\textbf{42.60}\pm \textbf{0.77\%}$}} &
			\multicolumn{4}{c|}{\multirow{1}{*}{$\textbf{62.84}\pm \textbf{0.98\%}$}} &
			\multicolumn{4}{c|}{\multirow{1}{*}{$\textbf{48.33}\pm \textbf{0.81\%}$}} &
			\multicolumn{4}{c}{\multirow{1}{*}{$\textbf{82.04}\pm \textbf{0.94\%}$}} \\
			\hline \hline
			\multicolumn{2}{l}{\multirow{9}{*}{5-shot}} & 
			\multicolumn{3}{l|}{\multirow{1}{*}{MAML~\cite{finn2017model}} } & 
			\multicolumn{4}{c|}{\multirow{1}{*}{$52.14\pm 0.65\%$}} &
			\multicolumn{4}{c|}{\multirow{1}{*}{$65.24\pm 0.77\%$}} &
			\multicolumn{4}{c|}{\multirow{1}{*}{$60.56\pm 0.74\%$}} &
			\multicolumn{4}{c}{\multirow{1}{*}{$88.55\pm 0.60\%$}} \\	 
			
			\multicolumn{2}{l}{\multirow{9}{*}{}} & 
			\multicolumn{3}{l|}{\multirow{1}{*}{Meta-Reg~\cite{yin2020meta}} } & 
			\multicolumn{4}{c|}{\multirow{1}{*}{$51.74\pm 0.68\%$}} &
			\multicolumn{4}{c|}{\multirow{1}{*}{$68.45\pm 0.81\%$}} &
			\multicolumn{4}{c|}{\multirow{1}{*}{$60.92\pm 0.69\%$}} &
			\multicolumn{4}{c}{\multirow{1}{*}{$89.08\pm 0.61\%$}} \\	
			
			\multicolumn{2}{l}{\multirow{9}{*}{}} & 
			\multicolumn{3}{l|}{\multirow{1}{*}{TAML~\cite{jamal2019task}} } & 
			\multicolumn{4}{c|}{\multirow{1}{*}{$52.75\pm 0.70\%$}} &
			\multicolumn{4}{c|}{\multirow{1}{*}{$66.09\pm 0.71\%$}} &
			\multicolumn{4}{c|}{\multirow{1}{*}{$61.14\pm 0.72\%$}} &
			\multicolumn{4}{c}{\multirow{1}{*}{$89.11\pm 0.59\%$}} \\	
			
			\multicolumn{2}{l}{\multirow{9}{*}{}} & 
			\multicolumn{3}{l|}{\multirow{1}{*}{Meta-Dropout~\cite{lee2019meta}}} & 
			\multicolumn{4}{c|}{\multirow{1}{*}{$52.53\pm 0.69\%$}} &
			\multicolumn{4}{c|}{\multirow{1}{*}{$67.32\pm 0.92\%$}} &
			\multicolumn{4}{c|}{\multirow{1}{*}{$60.86\pm 0.73\%$}} &
			\multicolumn{4}{c}{\multirow{1}{*}{$89.25\pm 0.59\%$}} \\	
			
			\multicolumn{2}{l}{\multirow{9}{*}{}} & 
			\multicolumn{3}{l|}{\multirow{1}{*}{MetaMix~\cite{yao2021improving}}} & 
			\multicolumn{4}{c|}{\multirow{1}{*}{$54.14\pm 0.73\%$}} &
			\multicolumn{4}{c|}{\multirow{1}{*}{$69.47\pm 0.60\%$}} &
			\multicolumn{4}{c|}{\multirow{1}{*}{$63.52\pm 0.73\%$}} &
			\multicolumn{4}{c}{\multirow{1}{*}{$89.75\pm 0.58\%$}} \\	
			
			\multicolumn{2}{l}{\multirow{9}{*}{}} & 
			\multicolumn{3}{l|}{\multirow{1}{*}{Meta-Maxup~\cite{ni2021data}} } & 
			\multicolumn{4}{c|}{\multirow{1}{*}{$53.02\pm 0.72\%$}} &
			\multicolumn{4}{c|}{\multirow{1}{*}{$69.16\pm 0.61\%$}} &
			\multicolumn{4}{c|}{\multirow{1}{*}{$62.64\pm 0.72\%$}} &
			\multicolumn{4}{c}{\multirow{1}{*}{$88.88\pm 0.57\%$}} \\	
			
			\multicolumn{2}{l}{\multirow{9}{*}{}} & 
			\multicolumn{3}{l|}{\multirow{1}{*}{MLTI~\cite{yao2021meta}} } & 
			\multicolumn{4}{c|}{\multirow{1}{*}{$55.22\pm 0.76\%$}} &
			\multicolumn{4}{c|}{\multirow{1}{*}{$70.69\pm 0.68\%$}} &
			\multicolumn{4}{c|}{\multirow{1}{*}{$64.55\pm 0.74\%$}} &
			\multicolumn{4}{c}{\multirow{1}{*}{$91.08\pm 0.54\%$}} \\	
			\cline{3-21}
			
			\multicolumn{2}{l}{\multirow{9}{*}{}} & 
			\multicolumn{3}{l|}{\multirow{1}{*}{TU }} & 
			\multicolumn{4}{c|}{\multirow{1}{*}{$56.33\pm 0.69\%$}} &
			\multicolumn{4}{c|}{\multirow{1}{*}{$73.97\pm 0.70\%$}} &
			\multicolumn{4}{c|}{\multirow{1}{*}{$64.81\pm 0.72\%$}} &
			\multicolumn{4}{c}{\multirow{1}{*}{$91.15\pm 0.60\%$}} \\

			\multicolumn{2}{l}{\multirow{9}{*}{}} & 
			\multicolumn{3}{l|}{\multirow{1}{*}{ATU }} & 
			\multicolumn{4}{c|}{\multirow{1}{*}{$\textbf{56.78}\pm \textbf{0.73\%}$}} &
			\multicolumn{4}{c|}{\multirow{1}{*}{$\textbf{74.50}\pm \textbf{0.90\%}$}} &
			\multicolumn{4}{c|}{\multirow{1}{*}{$\textbf{65.16}\pm \textbf{0.75\%}$}} &
			\multicolumn{4}{c}{\multirow{1}{*}{$\textbf{91.42}\pm \textbf{0.61\%}$}} \\
			\bottomrule[1pt]
		\end{tabular}
		}
\end{table}

\end{document}